\documentclass[10pt, twocolumn, letterpaper]{article}

\usepackage{iccv}              % To produce the CAMERA-READY version
\usepackage{multirow}
\usepackage{makecell}
\usepackage[T1]{fontenc}

\definecolor{iccvblue}{rgb}{0.21,0.49,0.74}
\definecolor{lightblue}{RGB}{173, 216, 230}
\definecolor{purpleblue}{RGB}{120, 100, 255}
\usepackage[pagebackref, breaklinks, colorlinks, allcolors=iccvblue]{hyperref}
\usepackage{caption}
\usepackage{graphicx}
\usepackage{xcolor}
\usepackage{bm}
\usepackage{tabularx}
\usepackage{array}
\usepackage{multirow}
\usepackage{booktabs}
\usepackage{natbib}
\definecolor{pink}{RGB}{255,182,193}
\definecolor{VeryLightPink}{RGB}{255, 248, 248}
\definecolor{VeryLightBlue}{RGB}{248, 248, 255}
\definecolor{VeryLightGreen}{RGB}{248, 255, 248}
\usepackage{float}

\makeatletter
\renewcommand{\paragraph}{\@startsection{paragraph}{4}{\z@}{1.2ex plus 0.5ex minus .2ex}{-1em}{\normalsize\bf}}
\makeatother

\def\paperID{9739} % *** Enter the Paper ID here
\def\confName{ICCV}
\def\confYear{2025}

\title{GS-Occ3D: Scaling Vision-only Occupancy Reconstruction\\
with Gaussian Splatting}

\author{
Baijun Ye\textsuperscript{1,2*},
Minghui Qin\textsuperscript{1*}, 
Saining Zhang\textsuperscript{3*}, 
Moonjun Goon\textsuperscript{1}, 
Shaoting Zhu\textsuperscript{1,2}, 
\\
Zebang Shen\textsuperscript{5}, 
Luan Zhang\textsuperscript{5},
Lu Zhang\textsuperscript{5},
Hao Zhao\textsuperscript{3,4},  
Hang Zhao\textsuperscript{1,2$\dagger$}
\\
\textsuperscript{1}IIIS, THU  \quad 
\textsuperscript{2}Shanghai Qi Zhi Institute \quad
\textsuperscript{3}AIR, THU \quad 
\textsuperscript{4}BAAI \quad
\textsuperscript{5}Mercedes-Benz Group China Ltd. \quad 
\\
}
\begin{document}
  \maketitle
{\let\thefootnote\relax\footnotetext{{$^*$ Equal contribution. Listing order is random.} $^\dagger$Corresponding author.}}
  \begin{abstract}

Occupancy is crucial for autonomous driving, providing essential geometric priors for perception and planning. However, existing methods predominantly rely on LiDAR-based occupancy annotations, which limits scalability and prevents leveraging vast amounts of potential crowdsourced data for auto-labeling. To address this, we propose GS-Occ3D, a scalable vision-only framework that directly reconstructs occupancy. Vision-only occupancy reconstruction poses significant challenges due to sparse viewpoints, dynamic scene elements, severe occlusions, and long-horizon motion. Existing vision-based methods primarily rely on mesh representation, which suffer from incomplete geometry and additional post-processing, limiting scalability. To overcome these issues, GS-Occ3D optimizes an explicit occupancy representation using an Octree-based Gaussian Surfel formulation, ensuring efficiency and scalability. Additionally, we decompose scenes into static background, ground, and dynamic objects, enabling tailored modeling strategies: (1) Ground is explicitly reconstructed as a dominant structural element, significantly improving large-area consistency; (2) Dynamic vehicles are separately modeled to better capture motion-related occupancy patterns. Extensive experiments on the Waymo dataset demonstrate that GS-Occ3D achieves state-of-the-art geometry reconstruction results. By curating vision-only binary occupancy labels from diverse urban scenes, we show their effectiveness for downstream occupancy models on Occ3D-Waymo and superior zero-shot generalization on Occ3D-nuScenes. It highlights the potential of large-scale vision-based occupancy reconstruction as a new paradigm for scalable auto-labeling. 
\href{https://gs-occ3d.github.io/}{Project Page.}
% Project Page:\url{https://gs-occ3d.github.io/}
\end{abstract}
  \section{Introduction}
\label{sec:intro}

\begin{figure}[tbp]
    \centering
    \includegraphics[width=\linewidth]{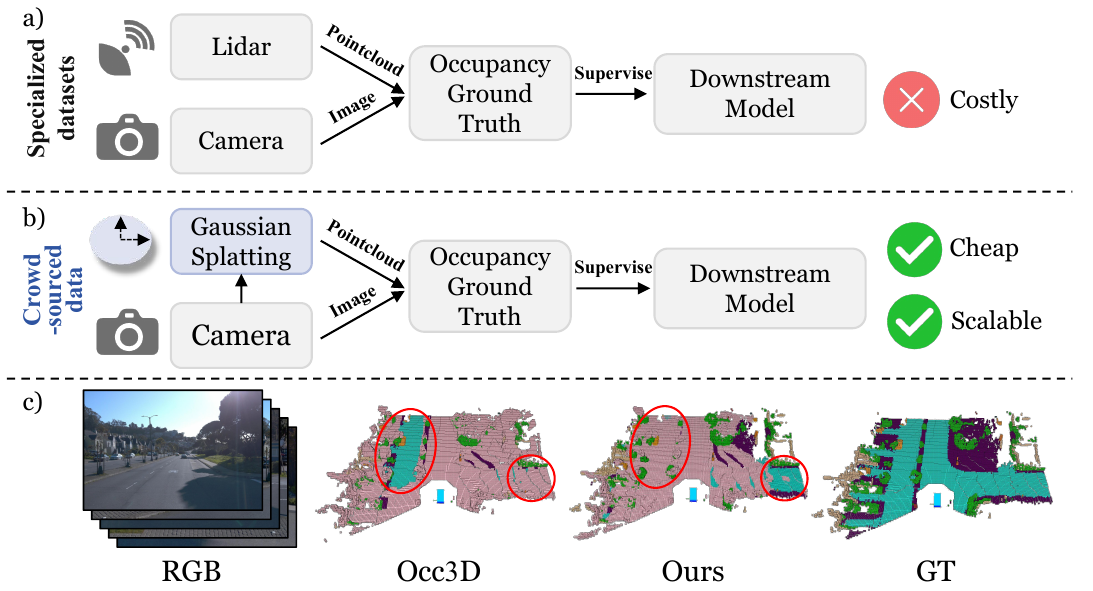}
    \vspace{-6mm} % too long(mhqin)
    \caption{\textbf{Overview of occupancy reconstruction pipelines.} a) Existing methods predominantly rely on LiDAR-based occupancy annotations, requiring \textit{costly specialized surveying vehicles}, which significantly limits scalability. b) In contrast, GS-Occ3D introduces a scalable, vision-only occupancy reconstruction framework that effectively harnesses \textit{abundant crowdsourced data from consumer-grade vehicles} for auto-labeling. Our approach enables affordable and scalable curation of high-quality occupancy labels. 
    c) 
    We present an overlay of the binary prediction \textcolor{pink}{(pink)} and the Occ3D-Waymo validation GT (other colors represent semantics), solely to visualize areas where the predictions are incomplete.
    Comparing models trained with two types of labels, we achieve generally comparable or even better \textbf{generalization geometry results} in certain setups.
    }
    \label{fig:teaser}
    \vspace{-6mm}
\end{figure}

% overall introduce
\noindent\textbf{Goal: Vision-only reconstruction.} Current occupancy reconstruction methods primarily depend on LiDAR-based annotations~\cite{huang2024gaussianformer,zheng2024gaussianad}. As shown in~\cref{fig:teaser}, these approaches face inherent scalability limitations due to the high costs associated with data collection and annotation. %Vision-centric data closed-loop frameworks in autonomous driving significantly enhance system scalability while reducing system-level operational costs. 
In contrast, vision-centric frameworks offer a promising alternative by leveraging large-scale crowdsourced data for self-supervised auto-labeling, eliminating the reliance on expensive LiDAR sensors \cite{tian2023occ3d,ma2024zoppframeworkzeroshotoffboard,wang2023openoccupancy,li2024sscbench,liu2024lidar}. Despite its potential, vision-only occupancy reconstruction remains a challenging task due to the sparsity of viewpoints, severe occlusions, dynamic scene elements, and long-horizon motion. For example, a typical scene in the Waymo dataset contains approximately 1,000 frames captured from only five cameras, resulting in limited co-visible regions across frames and incomplete scene observations.

% 1. how to get and how to utils geometry information
%sparse view hinder accurate geometry information
%Efficient acquisition and utilization of geometric information constitute the foundational basis for scalable occupancy reconstruction. In autonomous driving pipelines, the acquisition of precise geometric information typically requires expensive sensors. Although recent studies have demonstrated geometric priors through data-driven approaches \cite{wang2024dust3r,leroy2024mast3r,fan2024instantsplat,yin2023metric3d}, these methods fundamentally rely on LiDAR or high-precision RGB-D Ground Truth (GT) data. Consequently, vision-only systems face inherent limitations like texture-less region~\cite{chen2024pgsr,dai2024high,guedon2023sugar,yu2024gaussian} and struggle to obtain accurate geometry under sparse views along arbitrary trajectories such as stopping midway for traffic lights, long-distance travel and ego-static scenes. Specifically, a typical scene in Waymo consists of approximately 1000 frames captured by only five cameras, leading to limited co-visible regions between consecutive frames and incomplete observations of the entire scene.
%waymo
% metric scale？ ego-motion estimation and camera parameters. 

%scene representation

The success of scalable occupancy reconstruction hinges on an effective scene representation. Recent advances in scene representation learning, including Neural Radiance Fields (NeRF) \cite{mildenhall2021nerf,BarronMTHMS21,mueller2022instant,turki2022mega,tancik2022block,xu2023grid,ye2024blending} and 3D Gaussian Splatting (3DGS) \cite{kerbl20233d,lin2024vastgaussian,liu2024citygaussian,zhang2024drone,jiang2024horizongsunified3dgaussian,chen2024dogaussian,yang2024spectrallyprunedgaussianfields,kerbl2024hierarchical,yan2024street,zhou2024hugs,zhou2024hugsimrealtimephotorealisticclosedloop,chen2024omnire}, have shown remarkable rendering and novel view synthesis capabilities. Among these, 3DGS stands out for its efficiency and speed, but existing methods remain optimized for rendering quality rather than geometric accuracy, making them less suitable for occupancy reconstruction.

%mesh
\noindent\textbf{Limitations of prior methods}. (1) Prior works focusing on geometry reconstruction~\cite{chen2024pgsr,dai2024high,guedon2023sugar,yu2024gaussian,huang20242d} primarily target indoor, object-centric, or simple outdoor scenes. Extending these techniques to large-scale driving scenes leads to substantial accuracy degradation, particularly in weakly textured regions and along long-horizon, high-speed trajectories. (2) Existing methods that attempt street scene geometry reconstruction~\cite{song2025gvkf,lu2023urban,guo2023streetsurf,cui2024streetsurfgs} typically rely on mesh-based representations, which introduce overly smooth surfaces (in implicit methods) or fragmented, hole-ridden reconstruction (in GS-based methods). These issues necessitate extensive post-processing, hindering their scalability for efficient vision-only label curation. (3) Last but not least, most prior approaches \cite{lu2023urban,guo2023streetsurf,song2025gvkf,zhang2024rade,huang20242d,turkulainen2024dn,yu2024gsdf,huang20242d,yu2024gaussian} also focus solely on static scenes, failing to capture the dynamic aspects of real-world driving environments. This lack of dynamic object modeling reduces their applicability to downstream tasks~\cite{zeng2025futuresightdrive,song2025don,lin2025voteflowenforcinglocalrigidity,zheng2025world4driveendtoendautonomousdriving}. 
% that rely on complete dynamic occupancy reconstruction.

% make a conclusion for meaning and challenge
To summarize, scalable vision-only occupancy reconstruction is constrained by three major challenges: (1) Limited geometric priors from sparse views, making accurate reconstruction hard. (2) Degradation over long trajectories, leading to inconsistent and incomplete geometry. (3) Dynamic occlusions from moving objects, creating challenges in accurately modeling interactions in complex scenes.

To overcome these challenges, we introduce GS-Occ3D, a vision-only framework for scalable occupancy label curation that supports large-scale auto-labeling. Our method strategically decomposes driving environments into three geometrically distinct components: (1) Static backgrounds are modeled with an octree-based hierarchical surfel representation for multi-scale fidelity. (2) Ground surfaces are explicitly reconstructed as a dominant structural element to enhance large-area consistency. (3) Dynamic objects are processed separately to better capture motion-related occupancy patterns and reduce occlusion artifacts. 

By utilizing this tailored representation, we ensure high spatial-temporal consistency and preserve multi-scale geometric fidelity across long-horizon sequences. With this approach, we reconstruct the entire Waymo dataset to generate vision-only point clouds, which are then processed through a pipeline consisting of frame-wise division, multi-frame aggregation and voxelization. This enables the curation of vision-only occupancy labels, facilitating the training of state-of-the-art downstream occupancy models.

\noindent Overall, our contributions are summarized as follows:
\begin{itemize}
    \item We introduce a scalable pipeline for vision-only occupancy label curation, eliminating reliance on LiDAR while empowering downstream perception models.
    \item Our method effectively reconstructs the ground, background, and dynamic objects from panoramic street views along long trajectories. We outperform existing methods, even surpassing LiDAR-supervised baselines.
    \item We are the first to reconstruct the full Waymo dataset using a vision-only approach. We show the effectiveness of our labels for downstream occupancy models on Occ3D-Waymo and superior zero-shot generalization on Occ3D-nuScenes. This highlights the scalability of our approach for large-scale autonomous driving applications.
\end{itemize}

  \section{Related Works}
\noindent\textbf{Large-scale Scene Reconstruction.}
With the rapid advancement of NeRF \cite{mildenhall2021nerf,BarronMTHMS21,mueller2022instant,ye2024blending,yuan2024presight} and 3DGS \cite{kerbl20233d}, vision-based methods have revolutionized large-scale scene reconstruction.
NeRF-based methods segment scenes \cite{tancik2022block}, use grid-based ray association \cite{turki2022mega}, or integrate grids without decomposition \cite{xu2023grid}. GS-based approaches employ data partitioning for aerial training \cite{lin2024vastgaussian,liu2024citygaussian} or ensure global inference consistency \cite{chen2024dogaussian}. For static street scenes, Hierarchical-GS \cite{kerbl2024hierarchical} introduces Level-of-Detail (LOD), UC-GS \cite{zhang2024drone} refines car-view details via cross-view uncertainty, and Horizon-GS \cite{jiang2024horizongsunified3dgaussian} unifies aerial-street reconstruction with a coarse-to-fine LOD strategy.

For dynamic scenes, some works \cite{turki2023suds,yang2023emernerf} model all elements within a single field, while some works \cite{ost2021neural,yang2023unisim, wu2023mars, tonderski2024neurad} decompose scenes into foreground and background for better motion handling. Recent GS-based works \cite{yan2024street, zhou2023drivinggaussian, zhou2024hugs, zhou2024hugsimrealtimephotorealisticclosedloop, chen2024omnire, hess2024splatadrealtimelidarcamera,song2025coda,li2025mtgsmultitraversalgaussiansplatting,wang2025unifyingappearancecodesbilateral,xu2025cruisecooperativereconstructionediting} further improve the fidelity and efficiency.
More recently, some works \cite{li2024uniscene,lu2024infinicubeunboundedcontrollabledynamic,ye2025bevdiffuser,guo2025dist,yang2025x,nunes2024scaling} integrate diffusion models \cite{ho2020denoisingdiffusionprobabilisticmodels, esser2024scalingrectifiedflowtransformers} for diverse driving scene generation. However, they rely on LiDAR or prioritize rendering over geometry. We address this gap with a vision-only reconstruction method focused on geometry.

\noindent\textbf{Vision-only Geometry Reconstruction.}
Vision-only geometry reconstruction in urban scenes is challenging due to dynamic objects, sparse views, occlusion and long-horizon motion. NeRF-based methods like DNMP~\cite{lu2023urban} use neural mesh primitives, while StreetSurf~\cite{guo2023streetsurf} incorporates monocular cues (e.g., depth, normal) from pretrained models. However, their reliance on complete meshes or full-volume processing limits scalability.
GS-based methods advance object-centric and bounded scene reconstruction \cite{chen2024pgsr, dai2024high, guedon2023sugar, yu2024gaussian, zhang2024rade, huang20242d, turkulainen2024dn, yu2024gsdf}, but scaling to street scenes faces issues like uneven point distribution, holes, and floaters, making them unsuitable for downstream tasks. Recent grid-based improvements \cite{song2025gvkf, cui2024streetsurfgs} enhance mesh quality but struggle in weak-texture regions. In contrast, we reconstruct both ground and dynamic geometry, advancing scalable vision-only occupancy reconstruction and enabling seamless label curation for downstream tasks.

\noindent\textbf{3D Occupancy Prediction.}
3D occupancy prediction estimates the occupancy of each voxel in 3D space. With the rise of multiple benchmarks~\citep{tian2023occ3d, wang2023openoccupancy, li2024sscbench, liu2024lidar,wang2025uniocc} based on large-scale datasets~\citep{sun2020scalability, caesar2020nuscenes, behley2019semantickitti, liao2022kitti, chang2019argoverse}, this task has gained traction, especially for dynamic street-view applications.

Most methods rely on costly LiDAR-based occupancy labels.
LiDAR-based methods~\citep{song2017semantic,roldao2020lmscnet,li2025occmamba} complete scene occupancy from sparse LiDAR inputs, while camera-based methods predict 3D occupancy by extracting features from 2D images~\citep{cao2022monoscene, li2023voxformer, li2023bevstereo, li2024bevformer, zhang2023occformer, huang2023tri, li2023fb, ye2024cvt, pan2024co, liu2023bevfusion, wei2023surroundocc, park2022time,chen2025trackocc}. Monocular methods infer 3D structure from a single image via 2D-to-3D backprojection~\citep{cao2022monoscene} or depth-aware cross-attention~\citep{li2023voxformer}, while multi-view approaches generate 3D volumetric features from multiple camera views~\citep{li2023bevstereo, li2024bevformer, zhang2023occformer, huang2023tri, li2023fb, ye2024cvt, pan2024co, liu2023bevfusion, wei2023surroundocc}. Some works use GS to transform 2D images into dense gaussian representations~\citep{huang2024gaussianformer, zheng2024gaussianad}.
% Despite strong results, these methods heavily depend on manually annotated 3D labels
Despite strong results, they heavily depend on LiDAR-based annotations, which is costly and time-consuming. Recent vision approaches reduce reliance on LiDAR by using volume rendering with 2D supervision~\citep{huang2024selfocc, pan2024renderocc, zhang2023occnerf, park2022time, zheng2024veon, boeder2025gaussianflowocc,chambon2025gaussrender,liu2024let} or pretrained vision-language models~\citep{zhou2025occgs,boeder2024langocc, jiang2024gausstr}. Instead, we focus on leveraging vision-only 3D geometry labels, which is cheaper, more scalable and efficient for downstream occupancy models.

  \section{Methods}
\begin{figure*}[tbp]
    \centering
    \includegraphics[width=\linewidth]{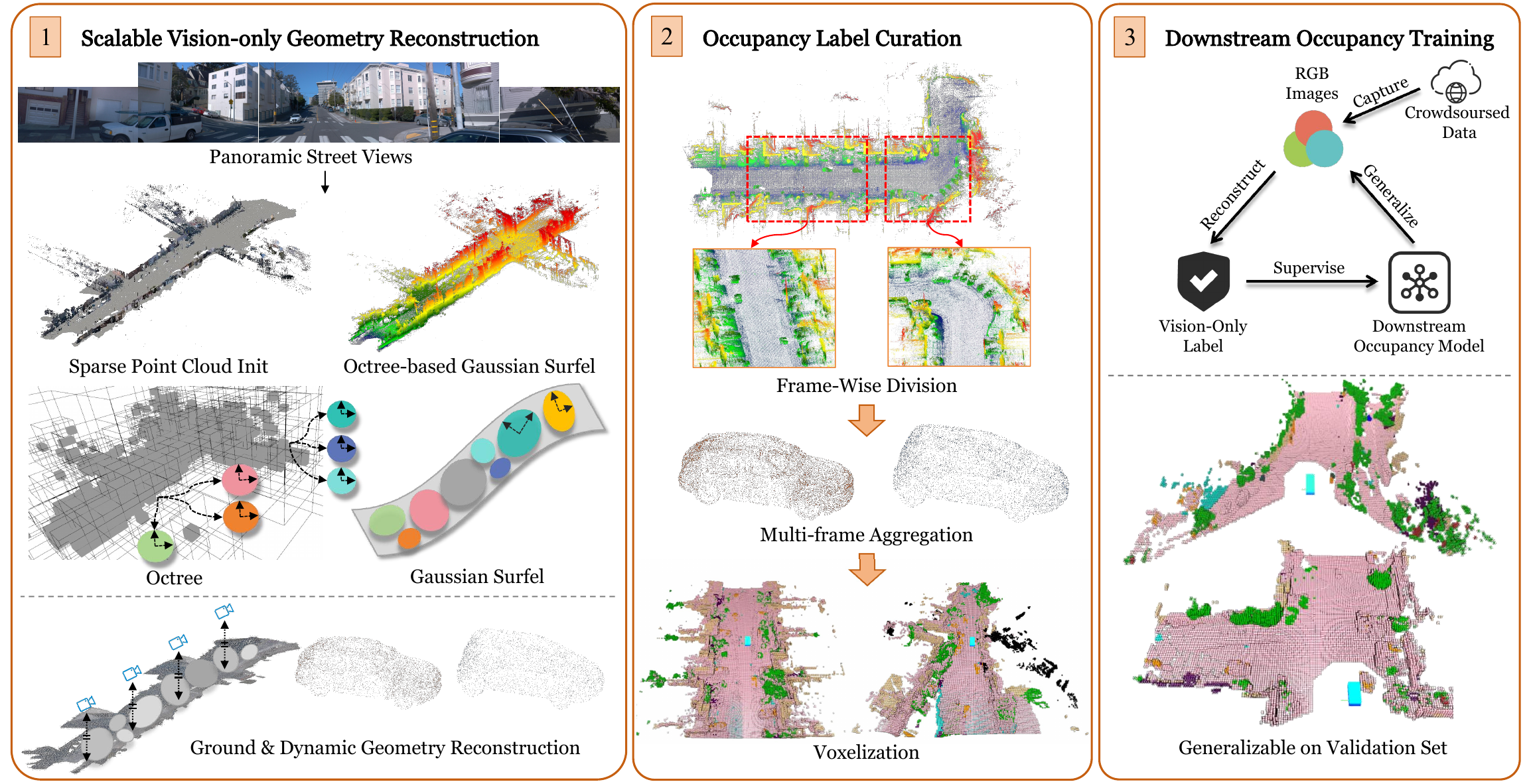}
    \vspace{-6mm}
    \caption{\textbf{Overview of the GS-Occ3D.} \textbf{Left:} Panoramic street
    views captured along long trajectories are used to generate a sparse point
    cloud and ground surfels as initialization. We adopt Octree-based Gaussian Surfel
    representation that integrates ground, background, and dynamic objects to achieve
    scalable vision-only geometry reconstruction. Here, we present an \textbf{uphill} scene with colors indicating height. \textbf{Middle:} Given the vision-only
    point cloud, our label curation pipeline applies frame-wise division
    and multi-frame aggregation to define appropriate perception ranges per
    frame, while increasing point cloud density, especially for dynamic objects
    with incomplete observations. Ray-casting is then applied to each frame to determine
    voxel occupancy, explicitly handling occlusions from the camera’s viewpoint.
    \textbf{Right:} The resulting vision-only labels can be used to train
    downstream occupancy models, enabling these models to generalize to unseen scenes
    with geometric reasoning capability. \textcolor{pink}{Pink} indicates the binary voxel, while other colors represent Occ3D labels.}
    \label{fig:system}
    \vspace{-5mm}
\end{figure*}

\cref{fig:system} illustrates the overview of our framework. 
We first generate a sparse point cloud and ground surfels from panoramic street views captured along long trajectories. For scalable vision-only geometry reconstruction, we use an octree-based Gaussian Surfel representation integrating ground, background, and dynamic objects.
Our label curation pipeline refines the vision-only point cloud through frame-wise division and multi-frame aggregation, increasing density, especially for dynamic objects. Ray-casting then resolves occlusions for accurate voxel occupancy labeling.
The resulting vision-only labels can supervise downstream occupancy models, improving their generalization and geometric reasoning capabilities.

\subsection{Scalable Vision-only Geometry Reconstruction}

\noindent\textbf{Preprocessing.}
 In the absence of geometric priors, we exclusively use detector-free SfM~\cite{sun2021loftr,he2023detectorfreestructuremotion} and ground gaussians detailed later to generate sparse point clouds as the initial scene representation. We also use an off-the-shelf segmentation model~\cite{cheng2021mask2former} for necessary decoupling.

\noindent\textbf{Octree-based Gaussian Surfel.}
To address the inherent lack of geometric priors in sparse views conditions, we adopt an octree-based gaussian surfel inspired by~\cite{ren2024octree}, using initial sparse point clouds as the scene skeleton to maximize the use of geometric information from the preprocessing stage. This structure enables hierarchical spatial partitioning and flexible adjustment during training, ensuring efficient, accurate, and scalable geometry reconstruction.

The dynamic octree structure adapts during training, expanding or contracting to match scene density and complexity.
Each sparse voxel can generate up to $m$ Gaussian primitives, confined to a small region centered at the voxel. These primitives serve as localized surfel representations, effectively approximating local surface geometry.

The voxel resolution adapts across different octree levels. Coarser levels feature lower spatial density, efficiently modeling global structures like walls and roads, while finer ones capture high-frequency details such as vegetation, buildings and object boundaries. The number of octree levels $K$ is determined by the observed distribution of distances between camera centers and input sparse point cloud, following a strategy similar to ~\cite{ren2024octree}.
It is computed as:
\begin{equation}
    K = \left\lfloor \log_{2}\left(\frac{d_{\max}}{d_{\min}}\right) \right\rceil
    + 1.
\end{equation}
where $d_{\min}$ and $d_{\max}$ are the smallest and largest distances between camera centers and SfM points. 
This ensures the octree provides sufficient granularity to capture both near- and far-field structures within the scene.

Once $K$ is determined, we initialize voxel centers at each octree level.
Starting with a base voxel size $\epsilon$ at the coarsest level, the voxel centers
at level $L$ are computed as:
\begin{equation}
    \mathbf{V}_{L}= \left\{ \left\lfloor\frac{\mathbf{P}}{\epsilon / 2^{L}}\right
    \rceil \cdot \frac{\epsilon}{2^{L}}\right\}.
\end{equation}
Here, $\mathbf{P}$ denotes the 3D coordinates of the sparse point cloud. This
hierarchical quantization keeps the voxel centers spatially aligned across
different levels, which is essential for efficient hierarchical geometry representation. Finally, after constructing the octree and initializing voxel centers, we assign
$m$ Gaussian surfels to each voxel.

% We use an adaptive anchor control strategy inspired by~\cite{lu2024scaffold,ren2024octree} to manage surfels. To enhance geometry completeness across scales, we apply cumulative LOD levels instead of a single one, to capture both coarse
% scene coverage and fine geometric details for high-fidelity reconstruction
% across varying spatial scales.
We use an adaptive anchor control strategy inspired by~\cite{lu2024scaffold,ren2024octree} to manage surfels. To enhance geometry completeness across scales, we apply cumulative LOD levels instead of a single one, to capture both coarse
scene coverage and fine geometric details for high-fidelity reconstruction.

\noindent\textbf{Ground Reconstruction.}
% Preliminary experiments show that existing geometry reconstruction methods, despite incorporating geometric constraints, struggle with weakly-textured ground regions in sparse-view street scenes, underscoring the need for a dedicated approach to enhance ground modeling accuracy.
Existing reconstruction methods, despite incorporating geometric constraints, struggle with weakly-textured ground in sparse-view scenes, highlighting the need for a dedicated ground modeling approach.

Assuming the road is parallel to the cameras, we initialize Ground Gaussian surfels by projecting camera poses onto the xy-plane, inspired by \cite{feng2024rogs}. To handle elevation, each surfel’s z-coordinate is adjusted using the nearest camera pose with a fixed height offset, while its orientation inherits the nearest camera’s rotation. This initialization aligns surfels with the road slope, enabling adaptation to various terrains as shown in \cref{fig:geometry}. Finally, planar regularization encourages smooth planar structures, further improving ground geometry as shown in~\cref{fig:road}.

\noindent
\textbf{Dynamic Reconstruction.} We assume each dynamic vehicle is associated with
a 3D bounding box predicted from RGB images and represented by points with tracked poses $\mathbf{R}_{t}$ and $\mathbf{t}_{t}$. We
initialize dynamic vehicles using vision-based 3D object tracking method~\cite{fischer2022cc3dt}. A fixed number of points are then sampled within each box.

To mitigate noise in the initial poses, we further refine them with learnable corrections,
following ~\cite{yan2024street}:
\begin{equation}
    \mathbf{R}_{t}' = \mathbf{R}_{t}\Delta \mathbf{R}_{t},\quad \mathbf{t}_{t}' =
    \mathbf{t}_{t}+ \Delta \mathbf{t}_{t},
\end{equation}
% \noindent where $\Delta \mathbf{t}_{t}$ is a learnable translation vector, and
% $\Delta \mathbf{R}_{t}$ is a rotation matrix constructed from a learnable yaw
% offset $\Delta \theta_{t}$. This design allows direct gradient computation,
% ensuring efficient training.
\noindent The learnable translation $\Delta \mathbf{t}_{t}$ and rotation $\Delta \mathbf{R}_{t}$ enable efficient training via direct gradient computation.

\noindent\textbf{Loss Function.}
% The total loss is a weighted sum of five components: RGB loss, geometry loss,
% object loss, ground loss, and sky loss, expressed as:
The total loss combines weighted RGB, geometry, object, ground, and sky components:
\begin{equation}
    \begin{split}
        L =&L_{rgb}+ \lambda_{geo}L_{geo}+ \lambda_{obj}L_{obj}\\&+ \lambda_{road}
        L_{road}+ \lambda_{sky}L_{sky},
    \end{split}
\end{equation}

\noindent where $\lambda_{geo}$, $\lambda_{obj}$, $\lambda_{road}$, and $\lambda_{sky}$ denote
the corresponding weights. The geometry loss $L_{geo}$ consists of three terms: surfel
regularization, depth distortion, and depth-normal consistency, formulated as:
\begin{equation}
    L_{geo}= \lambda_{s}L_{s}+ \lambda_{d}L_{d}+ \lambda_{n}L_{n}.
\end{equation}

% % where \(\lambda_{s}\), \(\lambda_{d}\), and \(\lambda_{n}\) are the corresponding weights.
Here, the RGB loss $L_{rgb}$ integrates L1 and D-SSIM losses to supervise RGB reconstruction, following \cite{kerbl20233d}.
The surfel regularization loss $L_{s}$ flattens Gaussians into surfels. The depth distortion loss $L_{d}$ and the normal depth consistency loss  $L_{n}$
encourage surfels to better conform to the geometry of the underlying scene, following the geometry constraints of \cite{huang20242d,yu2024gaussian}. Object loss $L_{obj}$ applies an entropy loss to the object opacity map, encouraging clearer decoupling between
foreground and background. Road smoothness loss $L_{road}$ preserves  flatness
by regularizing height variations between neighboring surfels. Sky loss
$L_{sky}$ applies a binary cross-entropy loss to the rendered opacity.

\subsection{Occupancy Labels Curation}

With scalable geometry reconstruction, it becomes feasible to curate vision-only
3D occupancy labels. However, the relatively sparse point cloud produced during
geometry reconstruction makes it hard to obtain dense voxel representations,
which leads to the sparsity problem. As the point cloud is densified, identifying occluded and invisible voxels from the camera views becomes essential. To address these issues, we employ frame-wise division and multi-frame aggregation to define per-frame perception ranges for vision-only point clouds and increase point-cloud density, particularly for dynamic objects with incomplete observations. Then we can operate ray-casting to determine the occupancy status of every voxel, explicitly handling occlusion.

\noindent\textbf{Frame-Wise Division.}
Unlike incremental LiDAR sequences, the reconstructed point cloud covers the entire scene in batches, which requires frame-wise division. To achieve this, we define a perception range centered on the camera pose, approximating the typical sensing range of LiDAR. Within this range, we uniformly sample points to form a single-sweep point cloud, ensuring that the number of points is consistent with that of a real LiDAR sweep.

\noindent\textbf{Multi-frame Aggregation.}
The reconstructed point cloud is relatively sparse, particularly for dynamic objects. To mitigate this, we aggregate points belonging to dynamic objects across frames, increasing their density.

Directly merging points across frames can cause smearing or distortion in dynamic objects like vehicles, so it is necessary to process them separately. Since our pipeline explicitly decouples static and dynamic components, we can extract optimized points within tracked bounding boxes without additional segmentation. These points are transformed from the sensor coordinate system to the box coordinate system, following a process similar to \cite{tian2023occ3d}. By concatenating these transformed points across frames, we effectively densify the point clouds for dynamic objects.

For static scenes, frame aggregation is unnecessary, as the static point cloud sequence is directly obtained by slicing the reconstructed scene. Unlike LiDAR-based labels, where frames provide complementary observations, static frames in our approach are inherently complete. After placing the densified dynamic points back into their corresponding bounding
boxes in each frame, we fuse the static scene with the aggregated dynamic
objects in the current frame, producing a dense single-frame point cloud.

\noindent\textbf{Voxelization.}
To generate a 3D occupancy grid from aggregated point clouds, a straightforward approach is to mark voxels containing points as occupied and others as free. However, due
to the limited camera field of view, some occupied voxels are only partially observed or entirely occluded from the camera's perspective, which can lead to incorrect
labeling as free. This ambiguity can confuse downstream models during training, making
it necessary to distinguish between free and unobserved voxels.
% Inspired by \cite{tian2023occ3d}, we use a ray-casting operation to determine the visibility of each voxel. Specifically, we trace a ray from the camera origin to
% the center of each occupied voxel. Along each ray, the first occupied voxel encountered is labeled as observed, while the rest are labeled as unobserved. Any voxel that is not traversed by any camera ray is also considered unobserved.
Inspired by \cite{tian2023occ3d}, we use ray-casting to determine voxel visibility. We cast a ray from the camera to each occupied voxel. Only the first occupied voxel encountered along any ray is labeled as observed, while all others are considered unobserved.

  \section{Experiments}

\begin{figure*}[tbp]
    \centering
    \includegraphics[width=\linewidth]{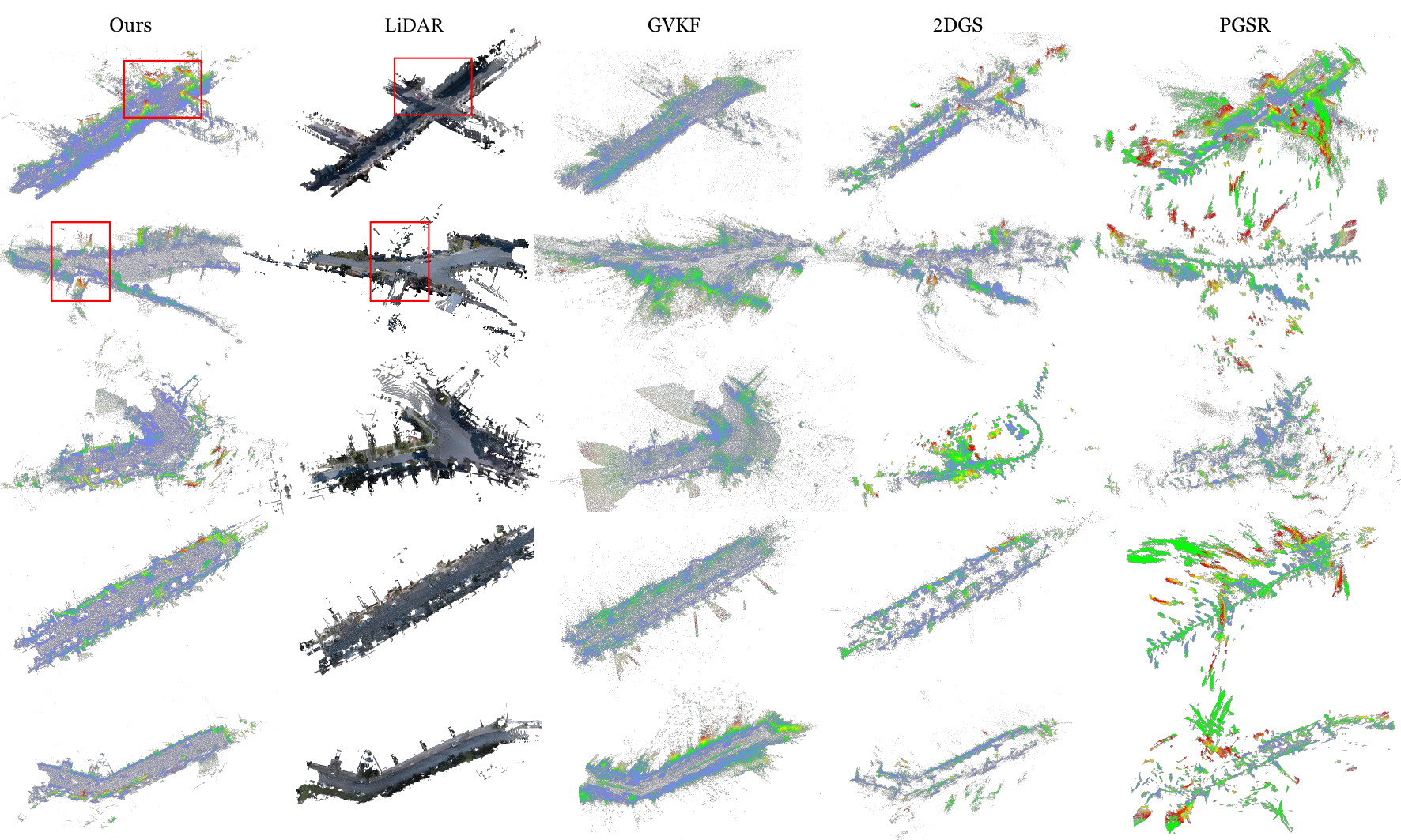}
    \vspace{-9mm}
    \caption{\textbf{Visualization of Geometry Reconstruction on Waymo.} The color represents CD with respect to LiDAR, ranging from \textcolor{purpleblue}{blue} (lower CD) to \textcolor{red}{red} (higher CD). We exhibit improved reconstruction fidelity in weakly-textured regions compared to other methods, while maintaining structural completeness comparable to LiDAR point cloud, even in the absence of geometric priors.}
    \label{fig:geometry}
    \vspace{-6mm}
\end{figure*}

\subsection{Geometry Reconstruction Experimental Setup}

Most geometry reconstruction methods focus on static scenes. For fairness, we compare our reconstructed static components with SOTA static scene methods.
Following \cite{yang2023emernerf}, we evaluate on the Waymo Static-32 split and select the most scalable method with high geometric accuracy to reconstruct the full Waymo dataset and curate labels.

\noindent\textbf{Datasets.} 
To evaluate performance and scalability in large-scale open scenes, we use the Waymo dataset \cite{sun2020scalability}. We use all five camera views and all frames in each scene, resulting in about 1,000 images per scene. LiDAR point clouds are used as reference to assess geometric accuracy.

\noindent\textbf{Baselines.} 
For geometry reconstruction, we evaluate our method against SOTA implicit methods (NeuS \cite{wang2021neus}, F2-NeRF \cite{wang2023f2}, StreetSurf \cite{guo2023streetsurf}) and GS-based methods (2DGS \cite{huang20242d}, PGSR \cite{chen2024pgsr}, GVKF \cite{song2025gvkf}). All explicit baselines use the same input point clouds for initialization.

\noindent\textbf{Metrics.} 
We evaluate reconstruction quality across geometry, rendering, and efficiency.  
For geometry, we measure the accuracy of both point clouds and meshes using Chamfer Distance (CD).  
For rendering quality, we report peak signal-to-noise ratio (PSNR).  
For efficiency, we record storage requirements, GPU memory usage, and training time.

\renewcommand{\thefootnote}{\fnsymbol{footnote}}

\subsection{Geometry Reconstruction Result Analysis}

\begin{figure}[tbp]
    \centering
    \includegraphics[width=0.95\linewidth]{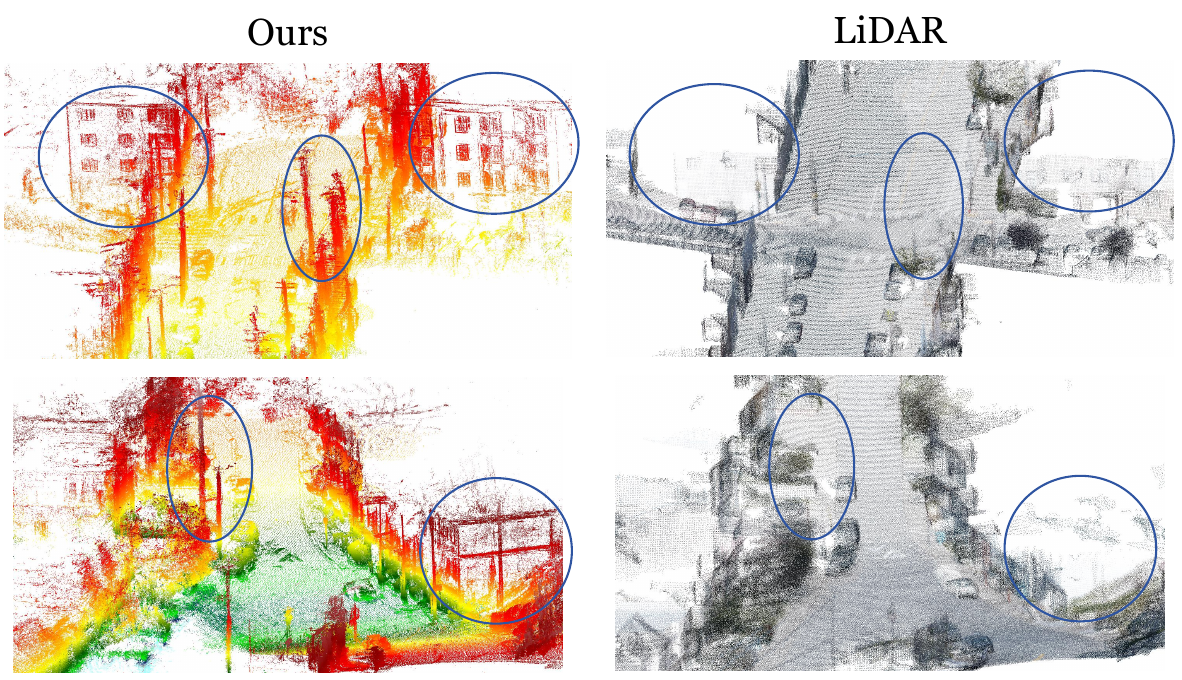}
    \vspace{-2mm}
    \caption{\textbf{Details of Geometry Reconstruction}. We present the detailed geometry of the \textcolor{red}{red-boxed} area in~\cref{fig:geometry}, achieving results that are both comparable and complementary to LiDAR. The first row is uphill, while the second is downhill followed by uphill.}
    \label{fig:detail}
    \vspace{-6mm}
    % \vspace{-6mm} %arxiv
\end{figure}

\noindent\textbf{SOTA Geometry Reconstruction.} 
\cref{tab:geometry} shows that we achieve SOTA geometry reconstruction, surpassing methods that use monocular cues or LiDAR-input or supervised like StreetSurf and NeuS, which demonstrates the reliability of geometry reconstruction. Besides geometry, we also maintain competitive rendering quality and training efficiency. This is partly due to the octree-based representation, which is both memory efficient and faster to train. Although we reconstruct denser point clouds, the structured octree helps accelerate geometry convergence while ensuring high rendering and geometric accuracy. Under sparse-view conditions, the octree-based gaussian surfel with LoD preserves geometric fidelity through multi-scale consistency.

\noindent\textbf{Comparable and Complementary to LiDAR.}
As shown in ~\cref{fig:geometry} and ~\cref{fig:detail}, our point cloud is globally comparable to LiDAR. In certain regions, such as tall buildings and thin poles, we achieve even higher quality, capturing finer details. Additionally, it provides more detailed textures and reconstructs areas beyond LiDAR’s limited range, making it reliable for downstream tasks. As observed in the comparison, our reasonable reconstruction beyond the LiDAR coverage is also counted as regions with high CD.

\begin{table}[tbp]
\vspace{2mm}
\centering
\small
\resizebox{\linewidth}{!}{
\renewcommand\tabcolsep{4pt} % Adjust column spacing
\begin{tabular}{c|c|c|c|ccc}
\toprule
% \textbf{Type}
\multirow{2}{*}{\rotatebox{90}{}} & \multirow{2}{*}{\textbf{Method}} & \multicolumn{1}{c|}{\textbf{Geometry}} & \textbf{Rendering} & \multicolumn{3}{c}{\textbf{Efficiency}} \\
& & CD $\downarrow$ & PSNR $\uparrow$ & MB $\downarrow$ & GB $\downarrow$ & Time $\downarrow$ \\
\midrule
\multirow{4}{*}{\rotatebox{90}{\textbf{Implicit}}} 
& NeuS\footnotemark[1] \cite{wang2021neus} & \underline{0.76} & 13.24 & 170 & 31 & 5.0h \\
& $\text{F}^2$-NeRF \cite{wang2023f2} & 886.77 & 24.70 & 130 & 24 & \textbf{0.8h} \\
& StreetSurf \cite{guo2023streetsurf} & 1.02 & \textbf{27.12} & 540 & 22 & 1.5h\\
& StreetSurf\footnotemark[2] \cite{guo2023streetsurf} & 0.90 & 26.85 & 245 & 21 & 1.5h \\
\midrule
\multirow{4}{*}{\rotatebox{90}{\textbf{Explicit}}} 
& PGSR\footnotemark[3] \cite{chen2024pgsr} & 2.90 & 22.61 & \textbf{78} & \textbf{4} & 1.5h \\
& 2DGS\footnotemark[3] \cite{huang20242d}  & 1.23 & 25.60 & 83 & 15 & \underline{1.0h} \\
& GVKF\footnotemark[3] \cite{song2025gvkf} & 0.82 & 25.87 & 65 & 24 & 2.0h \\
& Ours  & \textbf{0.56} & \underline{26.89} & \underline{80} & \underline{10} & \textbf{0.8h} \\
\bottomrule
\end{tabular}
}
\vspace{-2mm}
\caption{\textbf{Performance of implicit and explicit geometry reconstruction methods on the Waymo Static-32 Split}. NeuS\protect\footnotemark[1] uses 1 dense and 4 sparse LiDARs, StreetSurf\protect\footnotemark[2] uses 4 sparse LiDARs, and all other methods are vision-only.\protect\footnotemark[3] indicates using our ground gaussians. MB indicates storage size, GB indicates GPU memory, and Time indicates training time.
}
\label{tab:geometry}
\vspace{-7mm}
\end{table}

\renewcommand{\thefootnote}{\fnsymbol{footnote}} 

\begin{figure}[tbp]
    \centering
    \includegraphics[width=\linewidth]{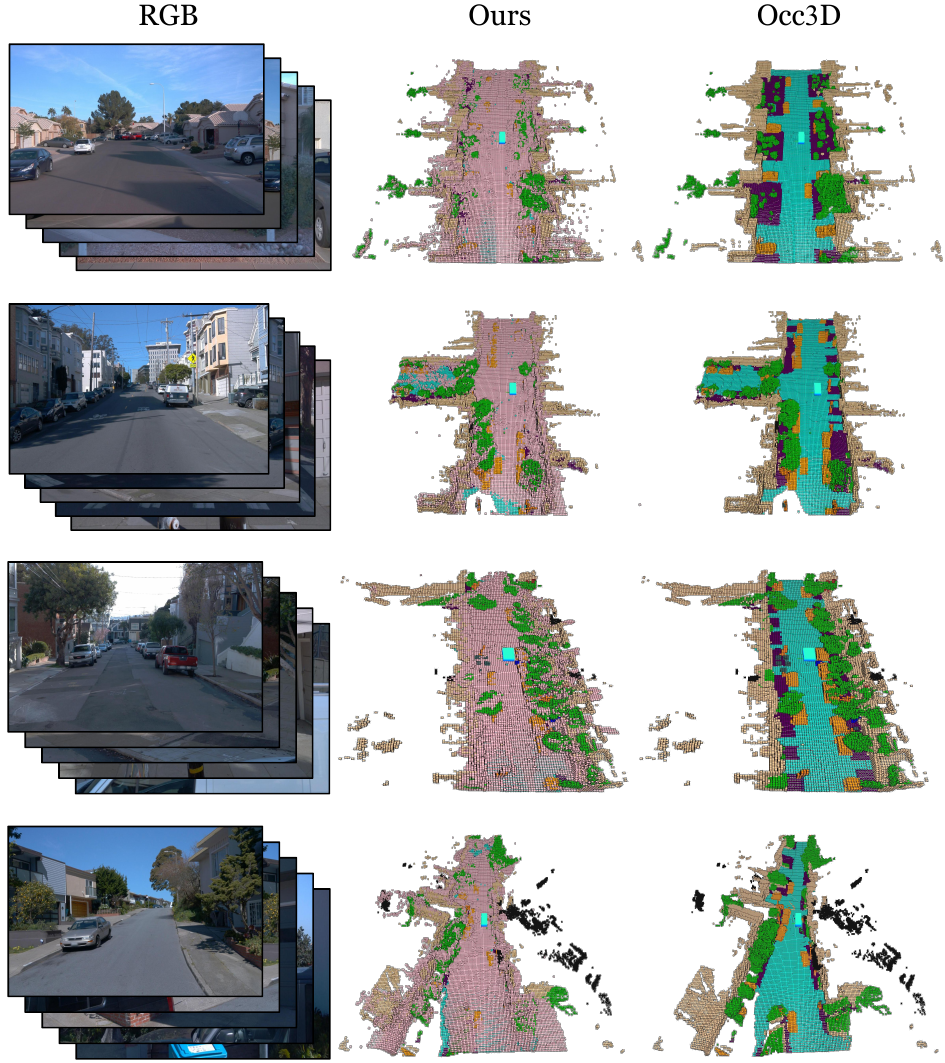}
    \vspace{-6mm}
    \caption{\textbf{Qualitative Results of Our Curated Labels.} We achieve globally comparable geometry to Occ3D, ensuring reliable supervision for occupancy model training without priors.}
% \textcolor{pink}{Pink} indicates the binary voxel, while other colors represent the Occ3D GT.
    \label{fig:voxel}
    % \vspace{-8mm} %arxiv
    \vspace{-6mm}
\end{figure}

\subsection{3D Occupancy Prediction Experimental Setup}
Through reconstructing the full Waymo dataset, we find vision-only methods fail to handle ego-static scenes. So we exclude these unreliable parts from the labels for training occupancy models. The experiments compare downstream occupancy models trained with the labels from Occ3D \cite{tian2023occ3d} and our labels. We evaluate the generalization performance on the validation set of Occ3D-Waymo and Occ3D-nuScenes and the fitting performance on the training set.

\noindent\textbf{Datasets.} Occ3D-Waymo is among the most diverse and comprehensively labeled open-source 3D occupancy datasets \cite{tian2023occ3d}, which contains 798 training scenes and 202 validation scenes, covering approximately 200,000 frames in total.
For fairness, we exclude ego-static scenes that cannot be reliably handled by vision-only reconstruction, resulting in 637 training scenes and 165 validation scenes.
We also use original Occ3D-nuScenes validation set including 150 scenes for evaluating zero-shot generalization.
% The spatial range is set to $[-40\ \text{m},\ 40\ \text{m}]$ for both x and y axes, and $[-1\ \text{m},\ 5.4\ \text{m}]$ for the z axis. The voxel grid size is $(0.4\ \text{m},\ 0.4\ \text{m},\ 0.4\ \text{m})$, yielding a resolution of $(200\times 200\times 16)$ for $(H, W, Z)$. 

\noindent\textbf{Baselines.} We compare our labels with LiDAR-based labels Occ3D \cite{tian2023occ3d} using SOTA occupancy model CVT-Occ \cite{ye2024cvt} which leverages geometric correspondences of 3D voxels over time to improve occupancy prediction accuracy.

\noindent\textbf{Metrics.} To evaluate the geometric accuracy and reliability of our labels, we use the Intersection over Union (IoU) to assess geometry performance of 3D occupancy prediction. For fair comparison, the evaluation only considers voxels within the visible region of the camera views.

\subsection{3D Occupancy Prediction Result Analysis}

\begin{table}[tbp]
\vspace{2mm} 
% \vspace{-2mm}
\centering
\small
\resizebox{\linewidth}{!}{
\setlength{\tabcolsep}{6pt}
\begin{tabular}{c|c|c c c c}
\toprule
\textbf{Train Labels} & \textbf{Eval Labels} & \textbf{IoU}~$\uparrow$ & \textbf{F1}~$\uparrow$ & \textbf{Prec.}~$\uparrow$ & \textbf{Rec.}~$\uparrow$ \\
\midrule
Ours (Waymo)   & \multirow{2}{*}{\makecell{Occ3D-Val(\textbf{Waymo})}} 
& 44.7 & 61.8 & 58.2 & 65.9 \\
Occ3D (Waymo)  & &\textbf{57.4} & \textbf{73.0} & \textbf{62.9} &  \textbf{87.0} \\
\midrule
Ours (Waymo)   & \multirow{2}{*}{\makecell{Occ3D-Val(\textbf{nuScenes})}}   & \textbf{33.4} & \textbf{50.1} & \textbf{62.5} & 41.8 \\
Occ3D (Waymo)  &                                        & 31.4          & 47.8          & 38.8          & \textbf{62.1} \\
\midrule
Ours (Waymo)   & \multirow{2}{*}{\makecell{Ours-Val(\textbf{Waymo})}}    & \textbf{46.8} & \textbf{63.8} & \textbf{54.6} & 76.6 \\
Occ3D (Waymo)  &                                        & 41.1         & 58.3          & 46.7          & \textbf{77.6} \\
\midrule
\multirow{2}{*}{Ours (Waymo) }   & \makecell{Ours-Train(\textbf{Waymo})}    & \textbf{50.6} & \textbf{67.2} & \textbf{54.3} & \textbf{88.1} \\
 &         \makecell{Occ3D-Train(\textbf{Waymo})}                                & 48.3 & 65.1 & 60.1 & 71.0 \\
\bottomrule
\end{tabular}
}
\vspace{-2mm}
\caption{\textbf{Generalization and Fitting Results on the Occ3D Dataset} for the SOTA occupancy model CVT-Occ \cite{ye2024cvt} under different training and evaluation label combinations.}
\label{tab:occ}
\vspace{-6mm}
\end{table}

\begin{figure}[tbp]
    \centering
    \vspace{-1.8mm}
    \includegraphics[width=\linewidth]{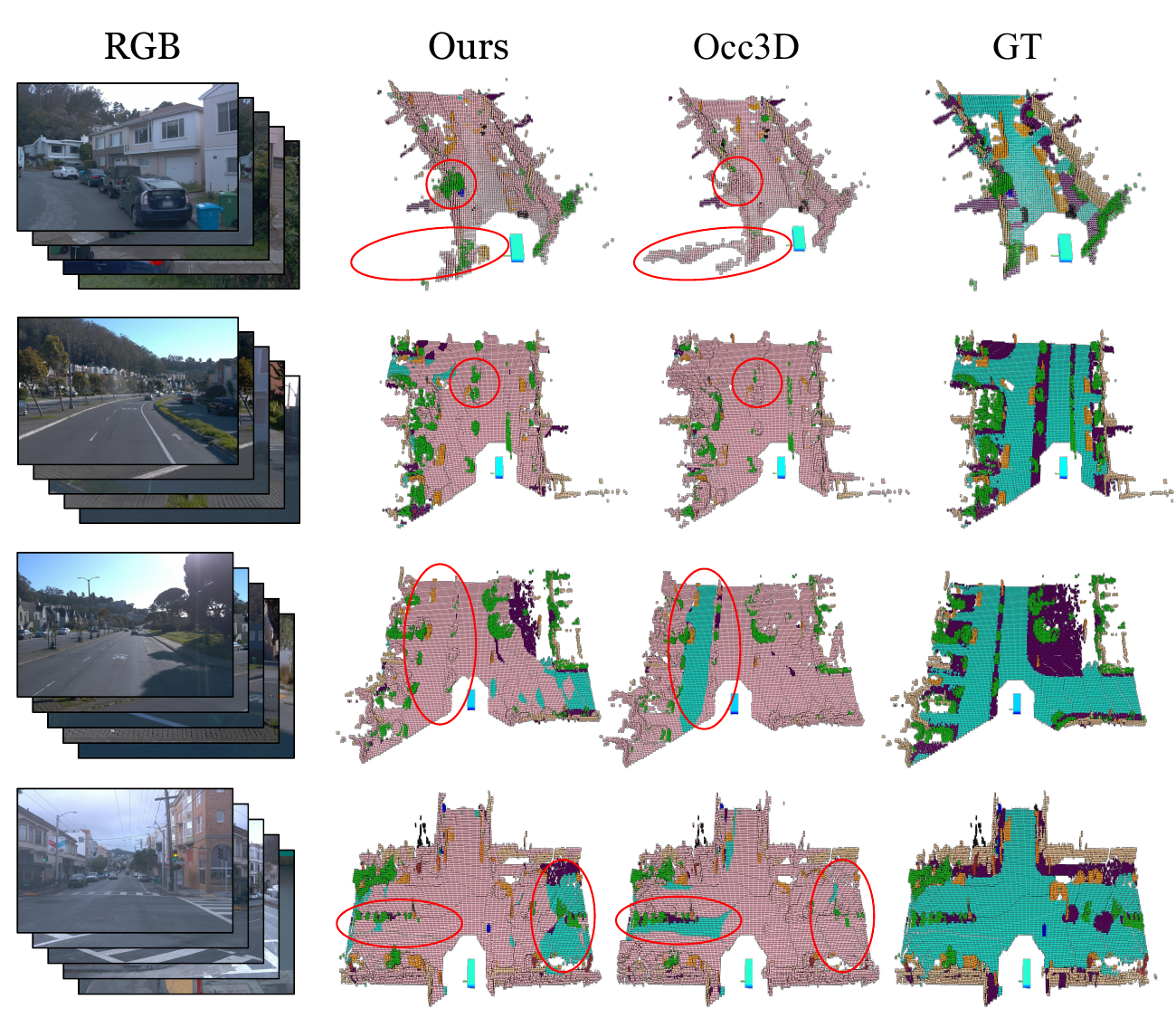}
    \vspace{-5mm}
    \caption{\textbf{Generalization Results on the Occ3D-Waymo Validation Set}. We evaluate the SOTA occupancy model CVT-Occ \cite{ye2024cvt} trained with our labels and Occ3D, achieving reasonable and overall comparable results.}
    \label{fig:Generalization}
    \vspace{-5mm}
\end{figure}

\noindent\textbf{Label Curation Results.} 
~\cref{fig:voxel} visualizes the comparison between our labels and Occ3D. Without geometric priors and initializing with sparse point clouds, we still achieve globally comparable geometry to the LiDAR-based labels. This ensures reliable geometry for downstream models.

\noindent\textbf{Comparable and Superior Zero-shot Occupancy Generalization Results.} 
~\cref{fig:Generalization} and ~\cref{tab:occ} show the generalization results of CVT-Occ \cite{ye2024cvt} on the validation set.
We demonstrate the ability to reconstruct generalizable geometry, achieving reasonable and overall comparable results on Occ3D-Waymo. 
Despite inherent camera limitations (e.g., Waymo's forward-facing views versus LiDAR's 360-degree coverage), we achieve slightly lower performance on the Occ3D-Val(Waymo) but still within a reasonable range. 
While both we and Occ3D perform better on our respective validation sets, we notably demonstrate superior zero-shot generalization on nuScenes with diverse camera settings, yielding more complete geometry in textured and distant regions. 
This is particularly impressive given that Occ3D relies on high-end LiDAR data. We rely purely on camera inputs yet delivers comparable or even better generalization performance in certain setups. This highlights not only the scalability of vision-only methods, but also their potential to match or surpass LiDAR-based baselines in real world.

\noindent\textbf{Occupancy Fitting Results.} 
% The last two rows of \cref{tab:occ} present the fitting results of CVT-Occ \cite{ye2024cvt} on different training sets. The performance gap evaluated on Occ3D-Train and Ours-Train is small, and the precision on the Occ3D-Train and Occ3D-Val remains similar. This indicates that our labels provide learnable geometry, enabling the model to capture meaningful geometry information.
As shown in the last two rows of \cref{tab:occ}, the performance gap evaluated on Occ3D-Train and Ours-Train is small, and the precision on the Occ3D-Train and Occ3D-Val remains similar. This indicates that our labels provide learnable geometry, enabling the model to capture meaningful geometry information.
% Our label provides learnable geometry, enabling the model to capture spatial hierarchies more effectively and distill meaningful structural relationships.

\begin{figure}[tbp]
    \centering
    % \vspace{-2mm}
    \includegraphics[width=\linewidth]{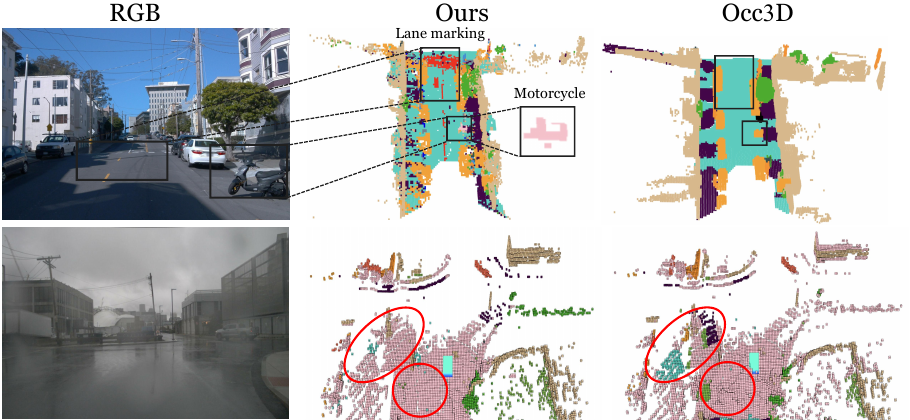}
    \vspace{-6mm}
  \caption{\textbf{More Visualization.} Up: Richer semantic labels. Down: superior generalization on Occ3D-nuScenes. \textcolor{pink}{Pink} indicates binary prediction, others show errors.}
    \label{fig:rebuttal}
    \vspace{-6mm} %arxiv
    % \vspace{-4mm}
\end{figure}
\noindent\textbf{Advantages over LiDAR.}
(1) \textbf{Wider coverage:} Vision-only reconstruction mitigates LiDAR's limited spatial range, excelling in large-scale areas and tall structures like high-rise buildings.
(2) \textbf{Superior zero-shot generalization:} Models trained with our labels can predict a wider range of geometries and generalize better to unseen scenes than LiDAR-based ones.
(3) \textbf{Cheap and Rich Semantics:} 
% Unlike LiDAR semantics which require labor-intensive annotation or costly multi-sensor fusion, vision provides rich and cheap semantics.
Unlike LiDAR semantics which require labor-intensive annotation or costly multi-sensor fusion, RGB images inherently capture color, texture, and object-class cues. 
Leveraging 2D segmentation models like ~\cite{cheng2021mask2former}, we produce a richer 3D category set (66 vs. Occ3D's 16), identifying objects LiDAR misses yet vital for driving like motorcycles, lane markings and crosswalks. (\cref{fig:rebuttal}).
% (4) \textbf{Greater potential in adverse weather:} Vision can leverage textures, semantics, and learned priors for more robust reconstruction in adverse weather. As shown in \cref{fig:rebuttal}, our method generalizes better than a LiDAR-based baseline in rainy conditions.
(4) \textbf{Greater potential in adverse weather:} Although adverse weather impacts both sensors, vision can use rich textures and semantics and learned priors from large-scale data, to reconstruct degraded scenes more effectively. From \cref{fig:rebuttal}, we even generalize better than LiDAR in rainy scenes.

\subsection{Ablation Studies}

% We conduct ablation studies to evaluate the contributions of different components in our framework.

\begin{figure}[tbp]
    \centering
    % \vspace{-2mm}
    \includegraphics[width=\linewidth]{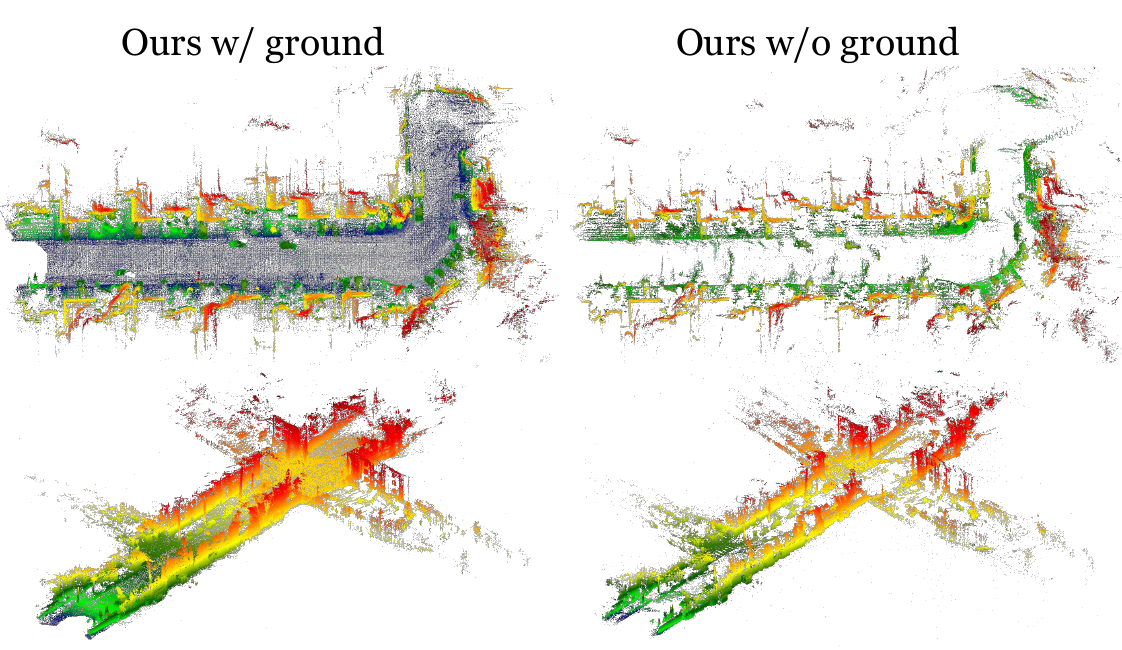}
    % \vspace{-6mm} %arxiv
    \vspace{-4mm}
  \caption{\textbf{Ablation Results of Ground Gaussians.} We show the effectiveness of our ground gaussians. Colors indicate height, ranging from \textcolor{blue}{blue} to \textcolor{red}{red}.}
    \label{fig:road}
    % \vspace{-5mm}%arxiv
    \vspace{-4mm}%cr
\end{figure}

\begin{table}[tbp]
% \vspace{2mm}%arxiv
\vspace{1mm}%cr
\centering
\small
\resizebox{\linewidth}{!}{ % 按照整个页面宽度缩放
\begin{tabular}{c|c|cc|c}
\toprule
\multirow{2}{*}{\textbf{Views}} & \multirow{2}{*}{\textbf{Method}} & \multicolumn{2}{c|}{\textbf{Geometry}} & \textbf{Rendering} \\
& & CD-Pcd $\downarrow$ & CD-Mesh $\downarrow$ & PSNR $\uparrow$ \\
\midrule
\multirow{4}{*}{\textbf{5Cam}} 
& PGSR \cite{chen2024pgsr}    & 3.63 & 4.41 & 19.18 \\
& 2DGS \cite{huang20242d}    & 1.25 & 2.14 & 23.42 \\
& GVKF \cite{song2025gvkf}    & 0.82 & \underline{1.22} & 25.87 \\
& Ours     & \textbf{0.56} & - & \underline{26.89} \\
\midrule
\multirow{4}{*}{\textbf{3Cam}} 
& PGSR \cite{chen2024pgsr}    & 2.90 & 3.03 & 22.61 \\
& 2DGS \cite{huang20242d}    & 1.23 & 1.85 & 25.60 \\
& GVKF \cite{song2025gvkf}    & 0.87 & \textbf{1.02} & 26.22 \\
& Ours     & \underline{0.66} & - & \textbf{26.96} \\
\bottomrule
\end{tabular}
}
\vspace{-2mm}
\caption{\textbf{Ablation Studies on Waymo Static-32 Split}. We evaluate varying camera counts and representations. All methods use our Ground Gaussians for fairness. Both chamfer distance of the point cloud and mesh are measured against LiDAR.}
\label{tab:cam}
% \vspace{-4mm} %arxiv
\vspace{-6mm}
\end{table}

\noindent\textbf{Input Camera Views.} 
% As shown in \cref{tab:cam}, compared to using only 3 cameras, we achieve better geometry reconstruction when utilizing the full panoramic 5-camera input. In contrast, some existing methods show reduced reconstruction and rendering quality with 5-camera input compared to 3-camera input due to the added geometry and rendering ambiguities introduced by forward-facing multi-view inputs in street scenes. 
As shown in \cref{tab:cam}, our method improves reconstruction with a 5-camera input over 3. In contrast, the performance degradation in other methods is primarily caused by the geometric and rendering ambiguities from forward-facing multi-view inputs.
However, our octree-based representation with LoD maintains structural fidelity and consistency across both global and local scales, enabling effective exploitation of multi-view observations. This makes our method well-suited for reconstructing diverse crowdsourced data and generating high-quality label.

\noindent\textbf{Choice on Point Cloud or Mesh.}
\cref{tab:cam} depicts that, for GS-based methods, directly representing geometry as a point cloud is preferable to mesh conversion. Mesh reconstruction introduces post-processing that transforms gaussians into surfaces, leading to information loss due to incomplete observations and the inherent limitations of meshing algorithms. This process often results in holes, sky enclosure artifacts, where the sky is misrepresented as a surface wrapping around the scene, and other errors that require further post-processing, which reduces scalability.

\noindent\textbf{Effectiveness of Our Ground Gaussians.} 
% \cref{fig:road} demonstrates the superiority of our ground gaussians in capturing detailed features of weakly-textured areas. Without separate handling of the ground, holes and abnormal protrusions can occur, distorting the geometry. 
\cref{fig:road} shows the effectiveness in capturing geometry of weakly-textured areas. Without separate handling of the ground, holes and abnormal protrusions can occur, distorting the geometry.

  \section{Conclusion}
GS-Occ3D is a scalable vision-only occupancy reconstruction framework that uses crowdsourced data from consumer vehicles for auto-labeling. It enables cost-efficient and scalable occupancy label curation. First, we adopt an Octree-based Gaussian Surfel formulation to reconstruct geometry for ground, background, and dynamic objects. Our vision-only reconstruction achieves SOTA geometric results. We then reconstruct the entire Waymo dataset, which covers diverse scenes. This enables us to validate the effectiveness of our labels for downstream occupancy models on Occ3D-Waymo and demonstrate superior zero-shot generalization on Occ3D-nuScenes. Our results highlight the potential of large-scale vision-based occupancy reconstruction as a new paradigm for scalable auto-labeling.
We hope this work inspires further research into using advanced reconstruction to empower vision-only label curation for downstream tasks.

% \noindent\textbf{Limitations.} Our method has several limitations.
% (1) Cameras inherently provide only front and side views, lacking rear view coverage, which leads to unavoidable information loss.  
% (2) Under nighttime conditions or exposure anomalies, the effective observation range of vision-based methods is significantly reduced. 
% (3) In ego-static scenarios, vision methods often fail to reconstruct geometry. Methods such as \cite{wang2024dust3r,leroy2024mast3r,fan2024instantsplat}, which pretrains with Waymo LiDAR point clouds, are incompatible with the vision-only setting. More recent methods such as \cite{wang2025vggt,chen2025long3rlongsequencestreaming} may overcome it.
% (4) We specifically focused on geometry reconstruction, consequently providing results solely on geometry generalization. Future work will aim to enhance both semantic and geometry reconstruction, particularly under challenging weather and lighting conditions, to achieve greater robustness in real-world applications.
\noindent\textbf{Limitations.} 
(1) Cameras inherently provide only front and side views, lacking rear view coverage, which leads to unavoidable information loss.  
(2)  Performance degrades at night or with exposure issues, which reduce the effective visual range.
(3) Vision methods often fail in ego-static scene. While methods pretrained on Waymo LiDAR~\cite{wang2024dust3r,leroy2024mast3r,fan2024instantsplat} are incompatible with our vision-only setting, recent works~\cite{wang2025vggt,chen2025long3rlongsequencestreaming} may offer solutions.
(4) We focus on geometry reconstruction, providing results solely on geometry generalization. Future work will aim to enhance both semantic and geometry reconstruction, particularly under challenging weather and lighting conditions, to achieve greater robustness in real-world applications.
  \section*{Acknowledgments}
% This work is supported by Tsinghua University - Mercedes Benz Joint Institute for Sustainable Mobility, the National Key R\&D Program of China (2022ZD0161700) and Tsinghua University Initiative Scientific Research Program.
This work is supported by the National Key Research and Development Program of China (2022YFB4701900), and the Tsinghua University - Mercedes Benz Joint Institute for Sustainable Mobility.
  { \small \bibliographystyle{ieeenat_fullname} \bibliography{main} }

\begin{thebibliography}{109}
\providecommand{\natexlab}[1]{#1}
\providecommand{\url}[1]{\texttt{#1}}
\expandafter\ifx\csname urlstyle\endcsname\relax
  \providecommand{\doi}[1]{doi: #1}\else
  \providecommand{\doi}{doi: \begingroup \urlstyle{rm}\Url}\fi

\bibitem[Barron et~al.(2021)Barron, Mildenhall, Tancik, Hedman, Martin{-}Brualla, and Srinivasan]{BarronMTHMS21}
Jonathan~T. Barron, Ben Mildenhall, Matthew Tancik, Peter Hedman, Ricardo Martin{-}Brualla, and Pratul~P. Srinivasan.
\newblock Mip-nerf: {A} multiscale representation for anti-aliasing neural radiance fields.
\newblock In \emph{ICCV}, 2021.

\bibitem[Behley et~al.(2019)Behley, Garbade, Milioto, Quenzel, Behnke, Stachniss, and Gall]{behley2019semantickitti}
Jens Behley, Martin Garbade, Andres Milioto, Jan Quenzel, Sven Behnke, Cyrill Stachniss, and Jurgen Gall.
\newblock Semantickitti: A dataset for semantic scene understanding of lidar sequences.
\newblock In \emph{Proceedings of the IEEE/CVF international conference on computer vision}, pages 9297--9307, 2019.

\bibitem[Boeder et~al.(2024)Boeder, Gigengack, and Risse]{boeder2024langocc}
Simon Boeder, Fabian Gigengack, and Benjamin Risse.
\newblock Langocc: Self-supervised open vocabulary occupancy estimation via volume rendering.
\newblock \emph{arXiv preprint arXiv:2407.17310}, 2024.

\bibitem[Boeder et~al.(2025)Boeder, Gigengack, and Risse]{boeder2025gaussianflowocc}
Simon Boeder, Fabian Gigengack, and Benjamin Risse.
\newblock Gaussianflowocc: Sparse and weakly supervised occupancy estimation using gaussian splatting and temporal flow.
\newblock \emph{arXiv preprint arXiv:2502.17288}, 2025.

\bibitem[Caesar et~al.(2020)Caesar, Bankiti, Lang, Vora, Liong, Xu, Krishnan, Pan, Baldan, and Beijbom]{caesar2020nuscenes}
Holger Caesar, Varun Bankiti, Alex~H Lang, Sourabh Vora, Venice~Erin Liong, Qiang Xu, Anush Krishnan, Yu Pan, Giancarlo Baldan, and Oscar Beijbom.
\newblock nuscenes: A multimodal dataset for autonomous driving.
\newblock In \emph{Proceedings of the IEEE/CVF conference on computer vision and pattern recognition}, pages 11621--11631, 2020.

\bibitem[Cao and De~Charette(2022)]{cao2022monoscene}
Anh-Quan Cao and Raoul De~Charette.
\newblock Monoscene: Monocular 3d semantic scene completion.
\newblock In \emph{Proceedings of the IEEE/CVF Conference on Computer Vision and Pattern Recognition}, pages 3991--4001, 2022.

\bibitem[Chambon et~al.(2025)Chambon, Zablocki, Boulch, Chen, and Cord]{chambon2025gaussrender}
Loïck Chambon, Eloi Zablocki, Alexandre Boulch, Mickaël Chen, and Matthieu Cord.
\newblock Gaussrender: Learning 3d occupancy with gaussian rendering, 2025.

\bibitem[Chang et~al.(2019)Chang, Lambert, Sangkloy, Singh, Bak, Hartnett, Wang, Carr, Lucey, Ramanan, et~al.]{chang2019argoverse}
Ming-Fang Chang, John Lambert, Patsorn Sangkloy, Jagjeet Singh, Slawomir Bak, Andrew Hartnett, De Wang, Peter Carr, Simon Lucey, Deva Ramanan, et~al.
\newblock Argoverse: 3d tracking and forecasting with rich maps.
\newblock In \emph{Proceedings of the IEEE/CVF conference on computer vision and pattern recognition}, pages 8748--8757, 2019.

\bibitem[Chen et~al.(2024{\natexlab{a}})Chen, Li, Ye, Wang, Xie, Zhai, Wang, Liu, Bao, and Zhang]{chen2024pgsr}
Danpeng Chen, Hai Li, Weicai Ye, Yifan Wang, Weijian Xie, Shangjin Zhai, Nan Wang, Haomin Liu, Hujun Bao, and Guofeng Zhang.
\newblock Pgsr: Planar-based gaussian splatting for efficient and high-fidelity surface reconstruction.
\newblock \emph{arXiv preprint arXiv:2406.06521}, 2024{\natexlab{a}}.

\bibitem[Chen and Lee(2024)]{chen2024dogaussian}
Yu Chen and Gim~Hee Lee.
\newblock Dogaussian: Distributed-oriented gaussian splatting for large-scale 3d reconstruction via gaussian consensus.
\newblock \emph{arXiv preprint arXiv:2405.13943}, 2024.

\bibitem[Chen et~al.(2024{\natexlab{b}})Chen, Yang, Huang, Lutio, Esturo, Ivanovic, Litany, Gojcic, Fidler, Pavone, Song, and Wang]{chen2024omnire}
Ziyu Chen, Jiawei Yang, Jiahui Huang, Riccardo~de Lutio, Janick~Martinez Esturo, Boris Ivanovic, Or Litany, Zan Gojcic, Sanja Fidler, Marco Pavone, Li Song, and Yue Wang.
\newblock Omnire: Omni urban scene reconstruction.
\newblock \emph{arXiv preprint arXiv:2408.16760}, 2024{\natexlab{b}}.

\bibitem[Chen et~al.(2025{\natexlab{a}})Chen, Li, Yang, Jiang, Li, and Zhao]{chen2025trackocc}
Zhuoguang Chen, Kenan Li, Xiuyu Yang, Tao Jiang, Yiming Li, and Hang Zhao.
\newblock Trackocc: Camera-based 4d panoptic occupancy tracking.
\newblock \emph{arXiv preprint arXiv:2503.08471}, 2025{\natexlab{a}}.

\bibitem[Chen et~al.(2025{\natexlab{b}})Chen, Qin, Yuan, Liu, and Zhao]{chen2025long3rlongsequencestreaming}
Zhuoguang Chen, Minghui Qin, Tianyuan Yuan, Zhe Liu, and Hang Zhao.
\newblock Long3r: Long sequence streaming 3d reconstruction, 2025{\natexlab{b}}.

\bibitem[Cheng et~al.(2021)Cheng, Choudhuri, Misra, Kirillov, Girdhar, and Schwing]{cheng2021mask2former}
Bowen Cheng, Anwesa Choudhuri, Ishan Misra, Alexander Kirillov, Rohit Girdhar, and Alexander~G Schwing.
\newblock Mask2former for video instance segmentation.
\newblock \emph{arXiv preprint arXiv:2112.10764}, 2021.

\bibitem[Cui et~al.(2024)Cui, Ye, Wang, Zhang, Zhou, and Li]{cui2024streetsurfgs}
Xiao Cui, Weicai Ye, Yifan Wang, Guofeng Zhang, Wengang Zhou, and Houqiang Li.
\newblock Streetsurfgs: Scalable urban street surface reconstruction with planar-based gaussian splatting.
\newblock \emph{arXiv preprint arXiv:2410.04354}, 2024.

\bibitem[Dai et~al.(2024)Dai, Xu, Xie, Liu, Wang, and Xu]{dai2024high}
Pinxuan Dai, Jiamin Xu, Wenxiang Xie, Xinguo Liu, Huamin Wang, and Weiwei Xu.
\newblock High-quality surface reconstruction using gaussian surfels.
\newblock In \emph{ACM SIGGRAPH 2024 Conference Papers}, pages 1--11, 2024.

\bibitem[Esser et~al.(2024)Esser, Kulal, Blattmann, Entezari, Müller, Saini, Levi, Lorenz, Sauer, Boesel, Podell, Dockhorn, English, Lacey, Goodwin, Marek, and Rombach]{esser2024scalingrectifiedflowtransformers}
Patrick Esser, Sumith Kulal, Andreas Blattmann, Rahim Entezari, Jonas Müller, Harry Saini, Yam Levi, Dominik Lorenz, Axel Sauer, Frederic Boesel, Dustin Podell, Tim Dockhorn, Zion English, Kyle Lacey, Alex Goodwin, Yannik Marek, and Robin Rombach.
\newblock Scaling rectified flow transformers for high-resolution image synthesis, 2024.

\bibitem[Fan et~al.(2024)Fan, Cong, Wen, Wang, Zhang, Ding, Xu, Ivanovic, Pavone, Pavlakos, et~al.]{fan2024instantsplat}
Zhiwen Fan, Wenyan Cong, Kairun Wen, Kevin Wang, Jian Zhang, Xinghao Ding, Danfei Xu, Boris Ivanovic, Marco Pavone, Georgios Pavlakos, et~al.
\newblock Instantsplat: Unbounded sparse-view pose-free gaussian splatting in 40 seconds.
\newblock \emph{arXiv preprint arXiv:2403.20309}, 2\penalty0 (3):\penalty0 4, 2024.

\bibitem[Feng et~al.(2024)Feng, Wu, Deng, and Wang]{feng2024rogs}
Zhiheng Feng, Wenhua Wu, Tianchen Deng, and Hesheng Wang.
\newblock Rogs: Large scale road surface reconstruction with meshgrid gaussian.
\newblock \emph{arXiv preprint arXiv:2405.14342}, 2024.

\bibitem[Fischer et~al.(2022)Fischer, Yang, Kumar, Sun, and Yu]{fischer2022cc3dt}
Tobias Fischer, Yung-Hsu Yang, Suryansh Kumar, Min Sun, and Fisher Yu.
\newblock Cc-3dt: Panoramic 3d object tracking via cross-camera fusion.
\newblock \emph{arXiv preprint arXiv:2212.01247}, 2022.

\bibitem[Gu{\'e}don and Lepetit(2023)]{guedon2023sugar}
Antoine Gu{\'e}don and Vincent Lepetit.
\newblock Sugar: Surface-aligned gaussian splatting for efficient 3d mesh reconstruction and high-quality mesh rendering.
\newblock \emph{arXiv preprint arXiv:2311.12775}, 2023.

\bibitem[Guo et~al.(2023)Guo, Deng, Li, Bai, Shi, Wang, Ding, Wang, and Li]{guo2023streetsurf}
Jianfei Guo, Nianchen Deng, Xinyang Li, Yeqi Bai, Botian Shi, Chiyu Wang, Chenjing Ding, Dongliang Wang, and Yikang Li.
\newblock Streetsurf: Extending multi-view implicit surface reconstruction to street views.
\newblock \emph{arXiv preprint arXiv:2306.04988}, 2023.

\bibitem[Guo et~al.(2025)Guo, Ding, Chen, Chen, Li, Zou, Lyu, Tan, Qi, Li, et~al.]{guo2025dist}
Jiazhe Guo, Yikang Ding, Xiwu Chen, Shuo Chen, Bohan Li, Yingshuang Zou, Xiaoyang Lyu, Feiyang Tan, Xiaojuan Qi, Zhiheng Li, et~al.
\newblock Dist-4d: Disentangled spatiotemporal diffusion with metric depth for 4d driving scene generation.
\newblock \emph{arXiv preprint arXiv:2503.15208}, 2025.

\bibitem[He et~al.(2023)He, Sun, Wang, Peng, Huang, Bao, and Zhou]{he2023detectorfreestructuremotion}
Xingyi He, Jiaming Sun, Yifan Wang, Sida Peng, Qixing Huang, Hujun Bao, and Xiaowei Zhou.
\newblock Detector-free structure from motion, 2023.

\bibitem[Hess et~al.(2024)Hess, Lindström, Fatemi, Petersson, and Svensson]{hess2024splatadrealtimelidarcamera}
Georg Hess, Carl Lindström, Maryam Fatemi, Christoffer Petersson, and Lennart Svensson.
\newblock Splatad: Real-time lidar and camera rendering with 3d gaussian splatting for autonomous driving, 2024.

\bibitem[Ho et~al.(2020)Ho, Jain, and Abbeel]{ho2020denoisingdiffusionprobabilisticmodels}
Jonathan Ho, Ajay Jain, and Pieter Abbeel.
\newblock Denoising diffusion probabilistic models, 2020.

\bibitem[Huang et~al.(2024{\natexlab{a}})Huang, Yu, Chen, Geiger, and Gao]{huang20242d}
Binbin Huang, Zehao Yu, Anpei Chen, Andreas Geiger, and Shenghua Gao.
\newblock 2d gaussian splatting for geometrically accurate radiance fields.
\newblock \emph{arXiv preprint arXiv:2403.17888}, 2024{\natexlab{a}}.

\bibitem[Huang et~al.(2023)Huang, Zheng, Zhang, Zhou, and Lu]{huang2023tri}
Yuanhui Huang, Wenzhao Zheng, Yunpeng Zhang, Jie Zhou, and Jiwen Lu.
\newblock Tri-perspective view for vision-based 3d semantic occupancy prediction.
\newblock In \emph{Proceedings of the IEEE/CVF conference on computer vision and pattern recognition}, pages 9223--9232, 2023.

\bibitem[Huang et~al.(2024{\natexlab{b}})Huang, Zheng, Zhang, Zhou, and Lu]{huang2024selfocc}
Yuanhui Huang, Wenzhao Zheng, Borui Zhang, Jie Zhou, and Jiwen Lu.
\newblock Selfocc: Self-supervised vision-based 3d occupancy prediction.
\newblock In \emph{Proceedings of the IEEE/CVF Conference on Computer Vision and Pattern Recognition}, pages 19946--19956, 2024{\natexlab{b}}.

\bibitem[Huang et~al.(2024{\natexlab{c}})Huang, Zheng, Zhang, Zhou, and Lu]{huang2024gaussianformer}
Yuanhui Huang, Wenzhao Zheng, Yunpeng Zhang, Jie Zhou, and Jiwen Lu.
\newblock Gaussianformer: Scene as gaussians for vision-based 3d semantic occupancy prediction.
\newblock In \emph{European Conference on Computer Vision}, pages 376--393. Springer, 2024{\natexlab{c}}.

\bibitem[Jiang et~al.(2024{\natexlab{a}})Jiang, Liu, Cheng, Wang, Lin, Su, Liu, and Wang]{jiang2024gausstr}
Haoyi Jiang, Liu Liu, Tianheng Cheng, Xinjie Wang, Tianwei Lin, Zhizhong Su, Wenyu Liu, and Xinggang Wang.
\newblock Gausstr: Foundation model-aligned gaussian transformer for self-supervised 3d spatial understanding.
\newblock \emph{arXiv preprint arXiv:2412.13193}, 2024{\natexlab{a}}.

\bibitem[Jiang et~al.(2024{\natexlab{b}})Jiang, Ren, Yu, Xu, Dong, Lu, Zhao, Lin, and Dai]{jiang2024horizongsunified3dgaussian}
Lihan Jiang, Kerui Ren, Mulin Yu, Linning Xu, Junting Dong, Tao Lu, Feng Zhao, Dahua Lin, and Bo Dai.
\newblock Horizon-gs: Unified 3d gaussian splatting for large-scale aerial-to-ground scenes, 2024{\natexlab{b}}.

\bibitem[Kerbl et~al.(2023)Kerbl, Kopanas, Leimk{\"u}hler, and Drettakis]{kerbl20233d}
Bernhard Kerbl, Georgios Kopanas, Thomas Leimk{\"u}hler, and George Drettakis.
\newblock 3d gaussian splatting for real-time radiance field rendering.
\newblock \emph{ACM Trans. Graph.}, 42\penalty0 (4):\penalty0 139--1, 2023.

\bibitem[Kerbl et~al.(2024)Kerbl, Meuleman, Kopanas, Wimmer, Lanvin, and Drettakis]{kerbl2024hierarchical}
Bernhard Kerbl, Andreas Meuleman, Georgios Kopanas, Michael Wimmer, Alexandre Lanvin, and George Drettakis.
\newblock A hierarchical 3d gaussian representation for real-time rendering of very large datasets.
\newblock \emph{ACM Transactions on Graphics (TOG)}, 43\penalty0 (4):\penalty0 1--15, 2024.

\bibitem[Leroy et~al.(2024)Leroy, Cabon, and Revaud]{leroy2024mast3r}
Vincent Leroy, Yohann Cabon, and J{\'e}r{\^o}me Revaud.
\newblock Grounding image matching in 3d with mast3r.
\newblock In \emph{European Conference on Computer Vision}, pages 71--91. Springer, 2024.

\bibitem[Li et~al.(2024{\natexlab{a}})Li, Guo, Liu, Zou, Ding, Chen, Zhu, Tan, Zhang, Wang, et~al.]{li2024uniscene}
Bohan Li, Jiazhe Guo, Hongsi Liu, Yingshuang Zou, Yikang Ding, Xiwu Chen, Hu Zhu, Feiyang Tan, Chi Zhang, Tiancai Wang, et~al.
\newblock Uniscene: Unified occupancy-centric driving scene generation.
\newblock \emph{arXiv preprint arXiv:2412.05435}, 2024{\natexlab{a}}.

\bibitem[Li et~al.(2025{\natexlab{a}})Li, Hou, Xing, Ma, Sun, and Zhang]{li2025occmamba}
Heng Li, Yuenan Hou, Xiaohan Xing, Yuexin Ma, Xiao Sun, and Yanyong Zhang.
\newblock Occmamba: Semantic occupancy prediction with state space models.
\newblock In \emph{Proceedings of the Computer Vision and Pattern Recognition Conference}, pages 11949--11959, 2025{\natexlab{a}}.

\bibitem[Li et~al.(2025{\natexlab{b}})Li, Qiu, Wu, Lindström, Su, Nießner, and Li]{li2025mtgsmultitraversalgaussiansplatting}
Tianyu Li, Yihang Qiu, Zhenhua Wu, Carl Lindström, Peng Su, Matthias Nießner, and Hongyang Li.
\newblock Mtgs: Multi-traversal gaussian splatting, 2025{\natexlab{b}}.

\bibitem[Li et~al.(2023{\natexlab{a}})Li, Bao, Ge, Yang, Sun, and Li]{li2023bevstereo}
Yinhao Li, Han Bao, Zheng Ge, Jinrong Yang, Jianjian Sun, and Zeming Li.
\newblock Bevstereo: Enhancing depth estimation in multi-view 3d object detection with temporal stereo.
\newblock In \emph{Proceedings of the AAAI Conference on Artificial Intelligence}, pages 1486--1494, 2023{\natexlab{a}}.

\bibitem[Li et~al.(2023{\natexlab{b}})Li, Yu, Choy, Xiao, Alvarez, Fidler, Feng, and Anandkumar]{li2023voxformer}
Yiming Li, Zhiding Yu, Christopher Choy, Chaowei Xiao, Jose~M Alvarez, Sanja Fidler, Chen Feng, and Anima Anandkumar.
\newblock Voxformer: Sparse voxel transformer for camera-based 3d semantic scene completion.
\newblock In \emph{Proceedings of the IEEE/CVF conference on computer vision and pattern recognition}, pages 9087--9098, 2023{\natexlab{b}}.

\bibitem[Li et~al.(2024{\natexlab{b}})Li, Li, Liu, Gong, Li, Chen, Wang, Li, Jiang, Yu, et~al.]{li2024sscbench}
Yiming Li, Sihang Li, Xinhao Liu, Moonjun Gong, Kenan Li, Nuo Chen, Zijun Wang, Zhiheng Li, Tao Jiang, Fisher Yu, et~al.
\newblock Sscbench: A large-scale 3d semantic scene completion benchmark for autonomous driving.
\newblock In \emph{2024 IEEE/RSJ International Conference on Intelligent Robots and Systems (IROS)}, pages 13333--13340. IEEE, 2024{\natexlab{b}}.

\bibitem[Li et~al.(2023{\natexlab{c}})Li, Yu, Austin, Fang, Lan, Kautz, and Alvarez]{li2023fb}
Zhiqi Li, Zhiding Yu, David Austin, Mingsheng Fang, Shiyi Lan, Jan Kautz, and Jose~M Alvarez.
\newblock Fb-occ: 3d occupancy prediction based on forward-backward view transformation.
\newblock \emph{arXiv preprint arXiv:2307.01492}, 2023{\natexlab{c}}.

\bibitem[Li et~al.(2024{\natexlab{c}})Li, Wang, Li, Xie, Sima, Lu, Yu, and Dai]{li2024bevformer}
Zhiqi Li, Wenhai Wang, Hongyang Li, Enze Xie, Chonghao Sima, Tong Lu, Qiao Yu, and Jifeng Dai.
\newblock Bevformer: learning bird's-eye-view representation from lidar-camera via spatiotemporal transformers.
\newblock \emph{IEEE Transactions on Pattern Analysis and Machine Intelligence}, 2024{\natexlab{c}}.

\bibitem[Liao et~al.(2022)Liao, Xie, and Geiger]{liao2022kitti}
Yiyi Liao, Jun Xie, and Andreas Geiger.
\newblock Kitti-360: A novel dataset and benchmarks for urban scene understanding in 2d and 3d.
\newblock \emph{IEEE Transactions on Pattern Analysis and Machine Intelligence}, 45\penalty0 (3):\penalty0 3292--3310, 2022.

\bibitem[Lin et~al.(2024)Lin, Li, Tang, Liu, Liu, Liu, Lu, Wu, Xu, Yan, et~al.]{lin2024vastgaussian}
Jiaqi Lin, Zhihao Li, Xiao Tang, Jianzhuang Liu, Shiyong Liu, Jiayue Liu, Yangdi Lu, Xiaofei Wu, Songcen Xu, Youliang Yan, et~al.
\newblock Vastgaussian: Vast 3d gaussians for large scene reconstruction.
\newblock \emph{arXiv preprint arXiv:2402.17427}, 2024.

\bibitem[Lin et~al.(2025)Lin, Wang, Nan, Kooij, and Caesar]{lin2025voteflowenforcinglocalrigidity}
Yancong Lin, Shiming Wang, Liangliang Nan, Julian Kooij, and Holger Caesar.
\newblock Voteflow: Enforcing local rigidity in self-supervised scene flow, 2025.

\bibitem[Liu et~al.(2024{\natexlab{a}})Liu, Gong, Fang, Xie, Li, Zhao, and Feng]{liu2024lidar}
Xinhao Liu, Moonjun Gong, Qi Fang, Haoyu Xie, Yiming Li, Hang Zhao, and Chen Feng.
\newblock Lidar-based 4d occupancy completion and forecasting.
\newblock In \emph{2024 IEEE/RSJ International Conference on Intelligent Robots and Systems (IROS)}, pages 11102--11109. IEEE, 2024{\natexlab{a}}.

\bibitem[Liu et~al.(2024{\natexlab{b}})Liu, Luo, Fan, Wang, Peng, and Zhang]{liu2024citygaussian}
Yang Liu, Chuanchen Luo, Lue Fan, Naiyan Wang, Junran Peng, and Zhaoxiang Zhang.
\newblock Citygaussian: Real-time high-quality large-scale scene rendering with gaussians.
\newblock In \emph{European Conference on Computer Vision}, pages 265--282. Springer, 2024{\natexlab{b}}.

\bibitem[Liu et~al.(2024{\natexlab{c}})Liu, Mou, Yu, Han, Mao, Xiong, and Wang]{liu2024let}
Yili Liu, Linzhan Mou, Xuan Yu, Chenrui Han, Sitong Mao, Rong Xiong, and Yue Wang.
\newblock Let occ flow: Self-supervised 3d occupancy flow prediction.
\newblock \emph{arXiv preprint arXiv:2407.07587}, 2024{\natexlab{c}}.

\bibitem[Liu et~al.(2023)Liu, Tang, Amini, Yang, Mao, Rus, and Han]{liu2023bevfusion}
Zhijian Liu, Haotian Tang, Alexander Amini, Xinyu Yang, Huizi Mao, Daniela~L Rus, and Song Han.
\newblock Bevfusion: Multi-task multi-sensor fusion with unified bird's-eye view representation.
\newblock In \emph{2023 IEEE international conference on robotics and automation (ICRA)}, pages 2774--2781. IEEE, 2023.

\bibitem[Lu et~al.(2023)Lu, Xu, Chen, Li, Lin, and Jiang]{lu2023urban}
Fan Lu, Yan Xu, Guang Chen, Hongsheng Li, Kwan-Yee Lin, and Changjun Jiang.
\newblock Urban radiance field representation with deformable neural mesh primitives.
\newblock In \emph{ICCV}, pages 465--476, 2023.

\bibitem[Lu et~al.(2024{\natexlab{a}})Lu, Yu, Xu, Xiangli, Wang, Lin, and Dai]{lu2024scaffold}
Tao Lu, Mulin Yu, Linning Xu, Yuanbo Xiangli, Limin Wang, Dahua Lin, and Bo Dai.
\newblock Scaffold-gs: Structured 3d gaussians for view-adaptive rendering.
\newblock In \emph{Proceedings of the IEEE/CVF Conference on Computer Vision and Pattern Recognition}, pages 20654--20664, 2024{\natexlab{a}}.

\bibitem[Lu et~al.(2024{\natexlab{b}})Lu, Ren, Yang, Shen, Wu, Gao, Wang, Chen, Chen, Fidler, and Huang]{lu2024infinicubeunboundedcontrollabledynamic}
Yifan Lu, Xuanchi Ren, Jiawei Yang, Tianchang Shen, Zhangjie Wu, Jun Gao, Yue Wang, Siheng Chen, Mike Chen, Sanja Fidler, and Jiahui Huang.
\newblock Infinicube: Unbounded and controllable dynamic 3d driving scene generation with world-guided video models, 2024{\natexlab{b}}.

\bibitem[Ma et~al.(2024)Ma, Zhou, Huang, Yang, Guo, Zhang, Dou, Qiao, Shi, and Li]{ma2024zoppframeworkzeroshotoffboard}
Tao Ma, Hongbin Zhou, Qiusheng Huang, Xuemeng Yang, Jianfei Guo, Bo Zhang, Min Dou, Yu Qiao, Botian Shi, and Hongsheng Li.
\newblock Zopp: A framework of zero-shot offboard panoptic perception for autonomous driving, 2024.

\bibitem[Mildenhall et~al.(2021)Mildenhall, Srinivasan, Tancik, Barron, Ramamoorthi, and Ng]{mildenhall2021nerf}
Ben Mildenhall, Pratul~P Srinivasan, Matthew Tancik, Jonathan~T Barron, Ravi Ramamoorthi, and Ren Ng.
\newblock Nerf: Representing scenes as neural radiance fields for view synthesis.
\newblock 65\penalty0 (1):\penalty0 99--106, 2021.

\bibitem[M\"uller et~al.(2022)M\"uller, Evans, Schied, and Keller]{mueller2022instant}
Thomas M\"uller, Alex Evans, Christoph Schied, and Alexander Keller.
\newblock Instant neural graphics primitives with a multiresolution hash encoding.
\newblock \emph{ACM TOG}, 2022.

\bibitem[Nunes et~al.(2024)Nunes, Marcuzzi, Mersch, Behley, and Stachniss]{nunes2024scaling}
Lucas Nunes, Rodrigo Marcuzzi, Benedikt Mersch, Jens Behley, and Cyrill Stachniss.
\newblock Scaling diffusion models to real-world 3d lidar scene completion.
\newblock In \emph{Proceedings of the IEEE/CVF Conference on Computer Vision and Pattern Recognition}, pages 14770--14780, 2024.

\bibitem[Ost et~al.(2021)Ost, Mannan, Thuerey, Knodt, and Heide]{ost2021neural}
Julian Ost, Fahim Mannan, Nils Thuerey, Julian Knodt, and Felix Heide.
\newblock Neural scene graphs for dynamic scenes.
\newblock In \emph{Proceedings of the IEEE/CVF Conference on Computer Vision and Pattern Recognition}, pages 2856--2865, 2021.

\bibitem[Pan et~al.(2024{\natexlab{a}})Pan, Wang, and Wang]{pan2024co}
Jingyi Pan, Zipeng Wang, and Lin Wang.
\newblock Co-occ: Coupling explicit feature fusion with volume rendering regularization for multi-modal 3d semantic occupancy prediction.
\newblock \emph{IEEE Robotics and Automation Letters}, 2024{\natexlab{a}}.

\bibitem[Pan et~al.(2024{\natexlab{b}})Pan, Liu, Zhang, Huang, Li, Xie, Wang, Liu, and Zhang]{pan2024renderocc}
Mingjie Pan, Jiaming Liu, Renrui Zhang, Peixiang Huang, Xiaoqi Li, Hongwei Xie, Bing Wang, Li Liu, and Shanghang Zhang.
\newblock Renderocc: Vision-centric 3d occupancy prediction with 2d rendering supervision.
\newblock In \emph{2024 IEEE International Conference on Robotics and Automation (ICRA)}, pages 12404--12411. IEEE, 2024{\natexlab{b}}.

\bibitem[Park et~al.(2022)Park, Xu, Yang, Keutzer, Kitani, Tomizuka, and Zhan]{park2022time}
Jinhyung Park, Chenfeng Xu, Shijia Yang, Kurt Keutzer, Kris Kitani, Masayoshi Tomizuka, and Wei Zhan.
\newblock Time will tell: New outlooks and a baseline for temporal multi-view 3d object detection.
\newblock \emph{arXiv preprint arXiv:2210.02443}, 2022.

\bibitem[Ren et~al.(2024)Ren, Jiang, Lu, Yu, Xu, Ni, and Dai]{ren2024octree}
Kerui Ren, Lihan Jiang, Tao Lu, Mulin Yu, Linning Xu, Zhangkai Ni, and Bo Dai.
\newblock Octree-gs: Towards consistent real-time rendering with lod-structured 3d gaussians.
\newblock \emph{arXiv preprint arXiv:2403.17898}, 2024.

\bibitem[Roldao et~al.(2020)Roldao, De~Charette, and Verroust-Blondet]{roldao2020lmscnet}
Luis Roldao, Raoul De~Charette, and Anne Verroust-Blondet.
\newblock Lmscnet: Lightweight multiscale 3d semantic completion.
\newblock In \emph{2020 International Conference on 3D Vision (3DV)}, pages 111--119. IEEE, 2020.

\bibitem[Song et~al.(2025{\natexlab{a}})Song, Cheng, and Wang]{song2025gvkf}
Gaochao Song, Chong Cheng, and Hao Wang.
\newblock Gvkf: Gaussian voxel kernel functions for highly efficient surface reconstruction in open scenes.
\newblock \emph{Advances in Neural Information Processing Systems}, 37:\penalty0 104792--104815, 2025{\natexlab{a}}.

\bibitem[Song et~al.(2025{\natexlab{b}})Song, Liang, Xia, Zimmer, Cao, Caesar, Festag, and Knoll]{song2025coda}
Rui Song, Chenwei Liang, Yan Xia, Walter Zimmer, Hu Cao, Holger Caesar, Andreas Festag, and Alois Knoll.
\newblock Coda-4dgs: Dynamic gaussian splatting with context and deformation awareness for autonomous driving.
\newblock \emph{arXiv preprint arXiv:2503.06744}, 2025{\natexlab{b}}.

\bibitem[Song et~al.(2017)Song, Yu, Zeng, Chang, Savva, and Funkhouser]{song2017semantic}
Shuran Song, Fisher Yu, Andy Zeng, Angel~X Chang, Manolis Savva, and Thomas Funkhouser.
\newblock Semantic scene completion from a single depth image.
\newblock In \emph{Proceedings of the IEEE conference on computer vision and pattern recognition}, pages 1746--1754, 2017.

\bibitem[Song et~al.(2025{\natexlab{c}})Song, Jia, Liu, Pan, Zhang, Wang, Zhang, Xu, Yang, and Luo]{song2025don}
Ziying Song, Caiyan Jia, Lin Liu, Hongyu Pan, Yongchang Zhang, Junming Wang, Xingyu Zhang, Shaoqing Xu, Lei Yang, and Yadan Luo.
\newblock Don't shake the wheel: Momentum-aware planning in end-to-end autonomous driving.
\newblock In \emph{Proceedings of the Computer Vision and Pattern Recognition Conference}, pages 22432--22441, 2025{\natexlab{c}}.

\bibitem[Sun et~al.(2021)Sun, Shen, Wang, Bao, and Zhou]{sun2021loftr}
Jiaming Sun, Zehong Shen, Yuang Wang, Hujun Bao, and Xiaowei Zhou.
\newblock Loftr: Detector-free local feature matching with transformers.
\newblock In \emph{Proceedings of the IEEE/CVF conference on computer vision and pattern recognition}, pages 8922--8931, 2021.

\bibitem[Sun et~al.(2020)Sun, Kretzschmar, Dotiwalla, Chouard, Patnaik, Tsui, Guo, Zhou, Chai, Caine, et~al.]{sun2020scalability}
Pei Sun, Henrik Kretzschmar, Xerxes Dotiwalla, Aurelien Chouard, Vijaysai Patnaik, Paul Tsui, James Guo, Yin Zhou, Yuning Chai, Benjamin Caine, et~al.
\newblock Scalability in perception for autonomous driving: Waymo open dataset.
\newblock In \emph{Proceedings of the IEEE/CVF conference on computer vision and pattern recognition}, pages 2446--2454, 2020.

\bibitem[Tancik et~al.(2022)Tancik, Casser, Yan, Pradhan, Mildenhall, Srinivasan, Barron, and Kretzschmar]{tancik2022block}
Matthew Tancik, Vincent Casser, Xinchen Yan, Sabeek Pradhan, Ben Mildenhall, Pratul~P Srinivasan, Jonathan~T Barron, and Henrik Kretzschmar.
\newblock Block-nerf: Scalable large scene neural view synthesis.
\newblock In \emph{CVPR}, pages 8248--8258, 2022.

\bibitem[Tian et~al.(2023)Tian, Jiang, Yun, Mao, Yang, Wang, Wang, and Zhao]{tian2023occ3d}
Xiaoyu Tian, Tao Jiang, Longfei Yun, Yucheng Mao, Huitong Yang, Yue Wang, Yilun Wang, and Hang Zhao.
\newblock Occ3d: A large-scale 3d occupancy prediction benchmark for autonomous driving.
\newblock \emph{Advances in Neural Information Processing Systems}, 36:\penalty0 64318--64330, 2023.

\bibitem[Tonderski et~al.(2024)Tonderski, Lindstr{\"o}m, Hess, Ljungbergh, Svensson, and Petersson]{tonderski2024neurad}
Adam Tonderski, Carl Lindstr{\"o}m, Georg Hess, William Ljungbergh, Lennart Svensson, and Christoffer Petersson.
\newblock Neurad: Neural rendering for autonomous driving.
\newblock In \emph{Proceedings of the IEEE/CVF Conference on Computer Vision and Pattern Recognition}, pages 14895--14904, 2024.

\bibitem[Turki et~al.(2022)Turki, Ramanan, and Satyanarayanan]{turki2022mega}
Haithem Turki, Deva Ramanan, and Mahadev Satyanarayanan.
\newblock Mega-nerf: Scalable construction of large-scale nerfs for virtual fly-throughs.
\newblock In \emph{CVPR}, pages 12922--12931, 2022.

\bibitem[Turki et~al.(2023)Turki, Zhang, Ferroni, and Ramanan]{turki2023suds}
Haithem Turki, Jason~Y Zhang, Francesco Ferroni, and Deva Ramanan.
\newblock Suds: Scalable urban dynamic scenes.
\newblock In \emph{Proceedings of the IEEE/CVF Conference on Computer Vision and Pattern Recognition}, pages 12375--12385, 2023.

\bibitem[Turkulainen et~al.(2024)Turkulainen, Ren, Melekhov, Seiskari, Rahtu, and Kannala]{turkulainen2024dn}
Matias Turkulainen, Xuqian Ren, Iaroslav Melekhov, Otto Seiskari, Esa Rahtu, and Juho Kannala.
\newblock Dn-splatter: Depth and normal priors for gaussian splatting and meshing.
\newblock \emph{arXiv preprint arXiv:2403.17822}, 2024.

\bibitem[Wang et~al.(2025{\natexlab{a}})Wang, Chen, Karaev, Vedaldi, Rupprecht, and Novotny]{wang2025vggt}
Jianyuan Wang, Minghao Chen, Nikita Karaev, Andrea Vedaldi, Christian Rupprecht, and David Novotny.
\newblock Vggt: Visual geometry grounded transformer.
\newblock In \emph{Proceedings of the Computer Vision and Pattern Recognition Conference}, pages 5294--5306, 2025{\natexlab{a}}.

\bibitem[Wang et~al.(2025{\natexlab{b}})Wang, Chen, Xiao, Xiao, Li, Chen, Ye, Xu, Zhang, Yan, Merriaux, Lei, Xue, and Zhao]{wang2025unifyingappearancecodesbilateral}
Nan Wang, Yuantao Chen, Lixing Xiao, Weiqing Xiao, Bohan Li, Zhaoxi Chen, Chongjie Ye, Shaocong Xu, Saining Zhang, Ziyang Yan, Pierre Merriaux, Lei Lei, Tianfan Xue, and Hao Zhao.
\newblock Unifying appearance codes and bilateral grids for driving scene gaussian splatting, 2025{\natexlab{b}}.

\bibitem[Wang et~al.(2021)Wang, Liu, Liu, Theobalt, Komura, and Wang]{wang2021neus}
Peng Wang, Lingjie Liu, Yuan Liu, Christian Theobalt, Taku Komura, and Wenping Wang.
\newblock Neus: Learning neural implicit surfaces by volume rendering for multi-view reconstruction.
\newblock \emph{arXiv preprint arXiv:2106.10689}, 2021.

\bibitem[Wang et~al.(2023{\natexlab{a}})Wang, Liu, Chen, Liu, Liu, Komura, Theobalt, and Wang]{wang2023f2}
Peng Wang, Yuan Liu, Zhaoxi Chen, Lingjie Liu, Ziwei Liu, Taku Komura, Christian Theobalt, and Wenping Wang.
\newblock F2-nerf: Fast neural radiance field training with free camera trajectories.
\newblock In \emph{Proceedings of the IEEE/CVF Conference on Computer Vision and Pattern Recognition}, pages 4150--4159, 2023{\natexlab{a}}.

\bibitem[Wang et~al.(2024)Wang, Leroy, Cabon, Chidlovskii, and Revaud]{wang2024dust3r}
Shuzhe Wang, Vincent Leroy, Yohann Cabon, Boris Chidlovskii, and Jerome Revaud.
\newblock Dust3r: Geometric 3d vision made easy.
\newblock In \emph{Proceedings of the IEEE/CVF Conference on Computer Vision and Pattern Recognition}, pages 20697--20709, 2024.

\bibitem[Wang et~al.(2023{\natexlab{b}})Wang, Zhu, Xu, Zhang, Wei, Chi, Ye, Du, Lu, and Wang]{wang2023openoccupancy}
Xiaofeng Wang, Zheng Zhu, Wenbo Xu, Yunpeng Zhang, Yi Wei, Xu Chi, Yun Ye, Dalong Du, Jiwen Lu, and Xingang Wang.
\newblock Openoccupancy: A large scale benchmark for surrounding semantic occupancy perception.
\newblock In \emph{Proceedings of the IEEE/CVF International Conference on Computer Vision}, pages 17850--17859, 2023{\natexlab{b}}.

\bibitem[Wang et~al.(2025{\natexlab{c}})Wang, Huang, Sun, Yan, Xing, Tu, and Li]{wang2025uniocc}
Yuping Wang, Xiangyu Huang, Xiaokang Sun, Mingxuan Yan, Shuo Xing, Zhengzhong Tu, and Jiachen Li.
\newblock Uniocc: A unified benchmark for occupancy forecasting and prediction in autonomous driving.
\newblock \emph{arXiv preprint arXiv:2503.24381}, 2025{\natexlab{c}}.

\bibitem[Wei et~al.(2023)Wei, Zhao, Zheng, Zhu, Zhou, and Lu]{wei2023surroundocc}
Yi Wei, Linqing Zhao, Wenzhao Zheng, Zheng Zhu, Jie Zhou, and Jiwen Lu.
\newblock Surroundocc: Multi-camera 3d occupancy prediction for autonomous driving.
\newblock In \emph{Proceedings of the IEEE/CVF International Conference on Computer Vision}, pages 21729--21740, 2023.

\bibitem[Wu et~al.(2023)Wu, Liu, Luo, Zhong, Chen, Xiao, Hou, Lou, Chen, Yang, Huang, Ye, Yan, Shi, Liao, and Zhao]{wu2023mars}
Zirui Wu, Tianyu Liu, Liyi Luo, Zhide Zhong, Jianteng Chen, Hongmin Xiao, Chao Hou, Haozhe Lou, Yuantao Chen, Runyi Yang, Yuxin Huang, Xiaoyu Ye, Zike Yan, Yongliang Shi, Yiyi Liao, and Hao Zhao.
\newblock Mars: An instance-aware, modular and realistic simulator for autonomous driving.
\newblock \emph{CICAI}, 2023.

\bibitem[Xu et~al.(2025)Xu, Zhang, Li, Ye, Chen, ang Gao, Zheng, Song, Peng, Miao, Jia, Shi, Yi, Zhao, Tang, Li, Yu, and Zhao]{xu2025cruisecooperativereconstructionediting}
Haoran Xu, Saining Zhang, Peishuo Li, Baijun Ye, Xiaoxue Chen, Huan ang Gao, Jv Zheng, Xiaowei Song, Ziqiao Peng, Run Miao, Jinrang Jia, Yifeng Shi, Guangqi Yi, Hang Zhao, Hao Tang, Hongyang Li, Kaicheng Yu, and Hao Zhao.
\newblock Cruise: Cooperative reconstruction and editing in v2x scenarios using gaussian splatting, 2025.

\bibitem[Xu et~al.(2023)Xu, Xiangli, Peng, Pan, Zhao, Theobalt, Dai, and Lin]{xu2023grid}
Linning Xu, Yuanbo Xiangli, Sida Peng, Xingang Pan, Nanxuan Zhao, Christian Theobalt, Bo Dai, and Dahua Lin.
\newblock Grid-guided neural radiance fields for large urban scenes.
\newblock In \emph{CVPR}, pages 8296--8306, 2023.

\bibitem[Yan et~al.(2024)Yan, Lin, Zhou, Wang, Sun, Zhan, Lang, Zhou, and Peng]{yan2024street}
Yunzhi Yan, Haotong Lin, Chenxu Zhou, Weijie Wang, Haiyang Sun, Kun Zhan, Xianpeng Lang, Xiaowei Zhou, and Sida Peng.
\newblock Street gaussians for modeling dynamic urban scenes.
\newblock \emph{arXiv preprint arXiv:2401.01339}, 2024.

\bibitem[Yang et~al.(2023{\natexlab{a}})Yang, Ivanovic, Litany, Weng, Kim, Li, Che, Xu, Fidler, Pavone, and Wang]{yang2023emernerf}
Jiawei Yang, Boris Ivanovic, Or Litany, Xinshuo Weng, Seung~Wook Kim, Boyi Li, Tong Che, Danfei Xu, Sanja Fidler, Marco Pavone, and Yue Wang.
\newblock Emernerf: Emergent spatial-temporal scene decomposition via self-supervision.
\newblock In \emph{International Conference on Learning Representations}, 2023{\natexlab{a}}.

\bibitem[Yang et~al.(2024)Yang, Zhu, Jiang, Ye, Chen, Zhang, Chen, Zhao, and Zhao]{yang2024spectrallyprunedgaussianfields}
Runyi Yang, Zhenxin Zhu, Zhou Jiang, Baijun Ye, Xiaoxue Chen, Yifei Zhang, Yuantao Chen, Jian Zhao, and Hao Zhao.
\newblock Spectrally pruned gaussian fields with neural compensation, 2024.

\bibitem[Yang et~al.(2025)Yang, Liang, Mei, Ma, Liu, and Lee]{yang2025x}
Yu Yang, Alan Liang, Jianbiao Mei, Yukai Ma, Yong Liu, and Gim~Hee Lee.
\newblock X-scene: Large-scale driving scene generation with high fidelity and flexible controllability.
\newblock \emph{arXiv preprint arXiv:2506.13558}, 2025.

\bibitem[Yang et~al.(2023{\natexlab{b}})Yang, Chen, Wang, Manivasagam, Ma, Yang, and Urtasun]{yang2023unisim}
Ze Yang, Yun Chen, Jingkang Wang, Sivabalan Manivasagam, Wei-Chiu Ma, Anqi~Joyce Yang, and Raquel Urtasun.
\newblock Unisim: A neural closed-loop sensor simulator.
\newblock In \emph{Proceedings of the IEEE/CVF Conference on Computer Vision and Pattern Recognition}, pages 1389--1399, 2023{\natexlab{b}}.

\bibitem[Ye et~al.(2024{\natexlab{a}})Ye, Liu, Ye, Chen, Wang, Yan, Shi, Zhao, and Zhou]{ye2024blending}
Baijun Ye, Caiyun Liu, Xiaoyu Ye, Yuantao Chen, Yuhai Wang, Zike Yan, Yongliang Shi, Hao Zhao, and Guyue Zhou.
\newblock Blending distributed nerfs with tri-stage robust pose optimization.
\newblock In \emph{2024 IEEE/RSJ International Conference on Intelligent Robots and Systems (IROS)}, pages 7975--7981. IEEE, 2024{\natexlab{a}}.

\bibitem[Ye et~al.(2025)Ye, Yaman, Cheng, Tao, Mallik, and Ren]{ye2025bevdiffuser}
Xin Ye, Burhaneddin Yaman, Sheng Cheng, Feng Tao, Abhirup Mallik, and Liu Ren.
\newblock Bevdiffuser: Plug-and-play diffusion model for bev denoising with ground-truth guidance.
\newblock In \emph{Proceedings of the Computer Vision and Pattern Recognition Conference}, pages 1495--1504, 2025.

\bibitem[Ye et~al.(2024{\natexlab{b}})Ye, Jiang, Xu, Li, and Zhao]{ye2024cvt}
Zhangchen Ye, Tao Jiang, Chenfeng Xu, Yiming Li, and Hang Zhao.
\newblock Cvt-occ: Cost volume temporal fusion for 3d occupancy prediction.
\newblock In \emph{European Conference on Computer Vision}, pages 381--397. Springer, 2024{\natexlab{b}}.

\bibitem[Yu et~al.(2024{\natexlab{a}})Yu, Lu, Xu, Jiang, Xiangli, and Dai]{yu2024gsdf}
Mulin Yu, Tao Lu, Linning Xu, Lihan Jiang, Yuanbo Xiangli, and Bo Dai.
\newblock Gsdf: 3dgs meets sdf for improved rendering and reconstruction.
\newblock \emph{arXiv preprint arXiv:2403.16964}, 2024{\natexlab{a}}.

\bibitem[Yu et~al.(2024{\natexlab{b}})Yu, Sattler, and Geiger]{yu2024gaussian}
Zehao Yu, Torsten Sattler, and Andreas Geiger.
\newblock Gaussian opacity fields: Efficient and compact surface reconstruction in unbounded scenes.
\newblock \emph{arXiv preprint arXiv:2404.10772}, 2024{\natexlab{b}}.

\bibitem[Yuan et~al.(2024)Yuan, Mao, Yang, Liu, Wang, and Zhao]{yuan2024presight}
Tianyuan Yuan, Yucheng Mao, Jiawei Yang, Yicheng Liu, Yue Wang, and Hang Zhao.
\newblock Presight: Enhancing autonomous vehicle perception with city-scale nerf priors.
\newblock In \emph{European Conference on Computer Vision}, pages 323--339. Springer, 2024.

\bibitem[Zeng et~al.(2025)Zeng, Chang, Xie, Liu, Bai, Pan, Xu, and Wei]{zeng2025futuresightdrive}
Shuang Zeng, Xinyuan Chang, Mengwei Xie, Xinran Liu, Yifan Bai, Zheng Pan, Mu Xu, and Xing Wei.
\newblock Futuresightdrive: Thinking visually with spatio-temporal cot for autonomous driving.
\newblock \emph{arXiv preprint arXiv:2505.17685}, 2025.

\bibitem[Zhang et~al.(2024{\natexlab{a}})Zhang, Fang, Shrestha, Liang, Long, and Tan]{zhang2024rade}
Baowen Zhang, Chuan Fang, Rakesh Shrestha, Yixun Liang, Xiaoxiao Long, and Ping Tan.
\newblock Rade-gs: Rasterizing depth in gaussian splatting.
\newblock \emph{arXiv preprint arXiv:2406.01467}, 2024{\natexlab{a}}.

\bibitem[Zhang et~al.(2023{\natexlab{a}})Zhang, Yan, Wei, Li, Liu, Tang, Duan, and Lu]{zhang2023occnerf}
Chubin Zhang, Juncheng Yan, Yi Wei, Jiaxin Li, Li Liu, Yansong Tang, Yueqi Duan, and Jiwen Lu.
\newblock Occnerf: Advancing 3d occupancy prediction in lidar-free environments.
\newblock \emph{arXiv preprint arXiv:2312.09243}, 2023{\natexlab{a}}.

\bibitem[Zhang et~al.(2024{\natexlab{b}})Zhang, Ye, Chen, Chen, Zhang, Peng, Shi, and Zhao]{zhang2024drone}
Saining Zhang, Baijun Ye, Xiaoxue Chen, Yuantao Chen, Zongzheng Zhang, Cheng Peng, Yongliang Shi, and Hao Zhao.
\newblock Drone-assisted road gaussian splatting with cross-view uncertainty.
\newblock \emph{arXiv preprint arXiv:2408.15242}, 2024{\natexlab{b}}.

\bibitem[Zhang et~al.(2023{\natexlab{b}})Zhang, Zhu, and Du]{zhang2023occformer}
Yunpeng Zhang, Zheng Zhu, and Dalong Du.
\newblock Occformer: Dual-path transformer for vision-based 3d semantic occupancy prediction.
\newblock In \emph{Proceedings of the IEEE/CVF International Conference on Computer Vision}, pages 9433--9443, 2023{\natexlab{b}}.

\bibitem[Zheng et~al.(2024{\natexlab{a}})Zheng, Tang, Wang, Wang, Ren, Feng, and Ma]{zheng2024veon}
Jilai Zheng, Pin Tang, Zhongdao Wang, Guoqing Wang, Xiangxuan Ren, Bailan Feng, and Chao Ma.
\newblock Veon: Vocabulary-enhanced occupancy prediction.
\newblock In \emph{European Conference on Computer Vision}, pages 92--108. Springer, 2024{\natexlab{a}}.

\bibitem[Zheng et~al.(2024{\natexlab{b}})Zheng, Wu, Zheng, Zuo, Xie, Yang, Pan, Hao, Jia, Lang, et~al.]{zheng2024gaussianad}
Wenzhao Zheng, Junjie Wu, Yao Zheng, Sicheng Zuo, Zixun Xie, Longchao Yang, Yong Pan, Zhihui Hao, Peng Jia, Xianpeng Lang, et~al.
\newblock Gaussianad: Gaussian-centric end-to-end autonomous driving.
\newblock \emph{arXiv preprint arXiv:2412.10371}, 2024{\natexlab{b}}.

\bibitem[Zheng et~al.(2025)Zheng, Yang, Xing, Zhang, Zheng, Gao, Li, Zhang, Xia, Jia, and Zhao]{zheng2025world4driveendtoendautonomousdriving}
Yupeng Zheng, Pengxuan Yang, Zebin Xing, Qichao Zhang, Yuhang Zheng, Yinfeng Gao, Pengfei Li, Teng Zhang, Zhongpu Xia, Peng Jia, and Dongbin Zhao.
\newblock World4drive: End-to-end autonomous driving via intention-aware physical latent world model, 2025.

\bibitem[Zhou et~al.(2024{\natexlab{a}})Zhou, Lin, Wang, Lu, Bai, Liu, Wang, Geiger, and Liao]{zhou2024hugsimrealtimephotorealisticclosedloop}
Hongyu Zhou, Longzhong Lin, Jiabao Wang, Yichong Lu, Dongfeng Bai, Bingbing Liu, Yue Wang, Andreas Geiger, and Yiyi Liao.
\newblock Hugsim: A real-time, photo-realistic and closed-loop simulator for autonomous driving, 2024{\natexlab{a}}.

\bibitem[Zhou et~al.(2024{\natexlab{b}})Zhou, Shao, Xu, Bai, Qiu, Liu, Wang, Geiger, and Liao]{zhou2024hugs}
Hongyu Zhou, Jiahao Shao, Lu Xu, Dongfeng Bai, Weichao Qiu, Bingbing Liu, Yue Wang, Andreas Geiger, and Yiyi Liao.
\newblock Hugs: Holistic urban 3d scene understanding via gaussian splatting.
\newblock \emph{arXiv preprint arXiv:2403.12722}, 2024{\natexlab{b}}.

\bibitem[Zhou et~al.(2023)Zhou, Lin, Shan, Wang, Sun, and Yang]{zhou2023drivinggaussian}
Xiaoyu Zhou, Zhiwei Lin, Xiaojun Shan, Yongtao Wang, Deqing Sun, and Ming-Hsuan Yang.
\newblock Drivinggaussian: Composite gaussian splatting for surrounding dynamic autonomous driving scenes.
\newblock \emph{arXiv preprint arXiv:2312.07920}, 2023.

\bibitem[Zhou et~al.(2025)Zhou, Wang, Wang, Wei, Dong, and Yang]{zhou2025occgs}
Xiaoyu Zhou, Jingqi Wang, Yongtao Wang, Yufei Wei, Nan Dong, and Ming-Hsuan Yang.
\newblock Occgs: Zero-shot 3d occupancy reconstruction with semantic and geometric-aware gaussian splatting.
\newblock \emph{arXiv preprint arXiv:2502.04981}, 2025.

\end{thebibliography}
\end{document}